\documentclass{article}
\usepackage{log_2023}						

\usepackage{booktabs}            
\usepackage{multirow}            
\usepackage{amsfonts}            
\usepackage{graphicx}            
\usepackage{duckuments}          
\usepackage[svgnames]{xcolor}
\usepackage{subfigure}
\usepackage{amsthm}

\theoremstyle{definition}
\newtheorem{definition}{Definition}

\usepackage[numbers,compress,sort]{natbib}

\title[Will More Expressive Graph Neural Networks do Better on Generative Tasks?]{Will More Expressive Graph Neural Networks do Better on Generative Tasks?}

\author[X. Zou et al.]{%
Xiandong Zou\\
Imperial College London, UK\\
\email{xz1320@ic.ac.uk}
\And
Xiangyu Zhao\\
Imperial College London, UK\\
\email{x.zhao22@imperial.ac.uk}
\AND
Pietro Li{\`o}\\
University of Cambridge, UK\\
\email{pietro.lio@cl.cam.ac.uk}
\And
Yiren Zhao\\
Imperial College London, UK\\
\email{a.zhao@imperial.ac.uk}
}

\begin{document}
\maketitle

\begin{abstract}
Graph generation poses a significant challenge as it involves predicting a complete graph with multiple nodes and edges based on simply a given label. This task also carries fundamental importance to numerous real-world applications, including de-novo drug and molecular design. In recent years, several successful methods have emerged in the field of graph generation. However, these approaches suffer from two significant shortcomings: (1) the underlying Graph Neural Network (GNN) architectures used in these methods are often underexplored; and (2) these methods are often evaluated on only a limited number of metrics. To fill this gap, we investigate the expressiveness of GNNs under the context of the molecular graph generation task, by replacing the underlying GNNs of graph generative models with more expressive GNNs. Specifically, we analyse the performance of six GNNs in two different generative frameworks---autoregressive generation models, such as GCPN and GraphAF, and one-shot generation models, such as GraphEBM---on six different molecular generative objectives on the ZINC-250k dataset. Through our extensive experiments, we demonstrate that advanced GNNs can indeed improve the performance of GCPN, GraphAF, and GraphEBM on molecular generation tasks, but GNN expressiveness is not a necessary condition for a good GNN-based generative model. Moreover, we show that GCPN and GraphAF with advanced GNNs can achieve state-of-the-art results across 17 other non-GNN-based graph generative approaches, such as variational autoencoders and Bayesian optimisation models, on the proposed molecular generative objectives (DRD2, Median1, Median2), which are important metrics for de-novo molecular design.
\end{abstract}

\section{Introduction}

Graph generation has always been viewed as a challenging task as it involves the prediction of a complete graph comprising multiple nodes and edges based on a given label. This task, however, holds paramount importance in a wide array of real-world applications, such as de-novo molecule design~\cite{drd2}. In de-novo molecule generation, the chemical space is discrete by nature, and the entire search space is huge, which is estimated to be between $10^{23}$ and $10^{60}$~\cite{chem}. The generation of novel and valid molecular graphs with desired physical, chemical and biological property objectives should also be considered, these together with the large search space makes it a difficult task, since \textit{these property objectives are highly complex and non-differentiable} \cite{GCPN}.

Recently, there has been significant progress in molecular graph generation with Graph Neural Network (GNN)-based deep generative models, such as the autoregressive generative models: Graph Convolutional Policy Network (GCPN)~\cite{GCPN} and Flow-based Autoregressive Model (GraphAF)~\cite{GraphAF} and the one-shot generative model: Graph Generation with Energy-based Model (GraphEBM)~\cite{ebm}. Those models all use the Relational Graph Convolutional Network (R-GCN)~\cite{RGCN} as their inner graph representation model. Meanwhile, several researchers have focused on improving GNN expressiveness by introducing architectural changes~\cite{GATv2, GSN, GearNet}, leading to the discovery of diverse forms of GNNs that excel in graph classification and regression tasks. Naturally, it is worth considering \emph{will these more expressive GNNs help in molecular graph generation, and will they perform better than the de-facto network used in GCPN, GraphAF, and GraphEBM?}

In addition, nowadays, there are many metrics used in de-novo molecule design (e.g. molecule's bioactivity against its corresponding disease targets: DRD2 and JNK3)~\cite{benchmark}. However, GCPN, GraphAF, and GraphEBM only consider two molecular generative objectives related to drug design: the Penalised logP and the QED. This brings two major drawbacks. Firstly, there is a lack of consideration of a broader set of generation metrics beyond those related to drug design. Secondly, both Penalised logP and the QED are incapable of effectively distinguishing between various generation models, rendering them disabled for such purpose~\cite{benchmark}.
 
In this work, we replace R-GCN in GCPN, GraphAF, and GraphEBM with more expressive GNNs. Then, we evaluate our proposed models: GCPN variants, GraphAF variants, and GraphEBM variants on a wide array of molecular generative objectives (e.g. DRD2, Median1, Median2), which are important metrics for de-novo molecular design, to derive more statistically significant results. The main contributions of our work are as follows:

\begin{itemize}
    \item Although GNN expressiveness defined by the Weisfeiler-Lehman (1-WL) graph isomorphism test works well on graph classification and regression tasks, it is not a necessary condition for a good GNN-based generative model. We observe empirically that the expressiveness of GNNs does not correlate well with their performance on GNN-based generative models, and GNNs incorporating edge feature extraction can improve GNN-based generative models.

    \item Although Penalised logP and QED are widely used in evaluating goal-directed graph generative models, they are not effective metrics to differentiate generative models. Other metrics, such as DRD2, Median1 and Median2, can better evaluate the ability of a graph generative model. 
    
    \item Our findings reveal that while there is no direct correlation between expressiveness and performance in graph generation, substituting the inner GNNs of GCPN, GraphAF, and GraphEBM with advanced GNNs like GearNet yields better performance (e.g. $102.51\%$ better in DRD2 and $48.96\%$ better in Median2). By doing so, these models surpass or reach comparable performance to state-of-the-art non-graph-based generative methods for de-novo molecule generation. 

\end{itemize}

\section{Related Work}
\label{rel}
A variety of deep generative models have been proposed for molecular graph generation recently~\cite{GCPN, GraphAF, ebm, SynNet,GA+D, GPBO, SMILES-VAE, Graph-MCTS, dataset1, DoG-Gen, MolDQN}. In this paper, we confine our scope to single-objective approaches for molecular generation. Specifically, our focus centres on the generation of organic molecules that possess a desired scalar metric, encompassing key physical, chemical, and biological properties.

\subsection{GNN-based Graph Generative Models}
\label{rel1}
Recently, there has been significant progress in molecular graph generation with GNN-based deep generative models, such as the autoregressive generation model (GCPN~\cite{GCPN} and GraphAF~\cite{GraphAF}) and the one-shot generation model (GraphEBM~\cite{ebm}). All of them use R-GCN~\cite{RGCN}, the state-of-the-art GNN at that time, as their inner graph representation model. Nevertheless, in recent years, the landscape has witnessed the emergence of increasingly expressive GNNs such as GATv2~\cite{GATv2}, GSN~\cite{GSN} and GearNet~\cite{GearNet}. These advanced GNN models have showcased superior performance in various tasks, including graph classification and regression, surpassing the capabilities of R-GCN. Furthermore, the methods GCPN, GraphAF, and GraphEBM only evaluate their performance in goal-directed molecule generation tasks using two commonly employed metrics in drug design: quantitative estimate of drug-likeness score (QED) and penalised octanol-water partition coefficient (Penalised logP).  
However, it is worth noting that many advanced graph generative approaches, as discussed in Section~\ref{ogga}, can achieve upper-limit results on benchmarks for QED and Penalised logP~\cite{benchmark, dataset1}. \textit{QED is likely to have a global maximum of 0.948 and even random sampling could reach that value. Penalised logP is unbounded and the relationship between Penalised logP values and molecular structures is fairly simple: adding carbons monotonically increases the estimated Penalised logP value~\cite{GCPN, dst, benchmark}.} This is presented in Figure \ref{d} in Appendix \ref{appb}. Consequently, one could argue that these scores have reached a saturation point, making them less meaningful as evaluation metrics. Only using these two metrics relevant to drug design on goal-directed graph generation tasks to assess the graph generative models is not convincing enough, and cannot provide insights for distinguishing different algorithms’ de-novo molecule generation ability~\cite {benchmark}.

\subsection{Non-GNN-based Graph Generative Models}
\label{ogga}
There exist also non-GNN-based graph generative models. Genetic Algorithms (GA) are generation approaches by relying on biologically inspired operators such as mutation, crossover and selection. Bayesian Optimisation (BO)~\cite{BO} is an approach that uses a sequential optimisation technique that leverages probabilistic models to search for the optimal solution. Variational Autoencoders (VAEs)~\cite{VAE} is a type of generative model in machine learning that combines elements of both autoencoders and probabilistic latent variable models to learn and generate data by mapping it to a latent space with continuous distributions. Monte-Carlo Tree Search (MCTS) constructs a search tree by iteratively selecting actions, simulating possible outcomes, and propagating the results to inform future decisions, ultimately aiming to find the optimal solution. Hill Climbing (HC) is an iterative optimisation algorithm that starts with an arbitrary solution to a problem, and then attempts to find a better solution by making an incremental change to the solution. Reinforcement Learning (RL) learns how intelligent agents take actions in an environment to maximise the cumulative reward by transitioning through different states. Details about the non-GNN-based graph generative models we use as baselines in Section \ref{eva} can be found in Appendix \ref{ngnn}.
\section{Method}
\label{Method}
\begin{figure}[t]
\centering 
\includegraphics[width=\textwidth]{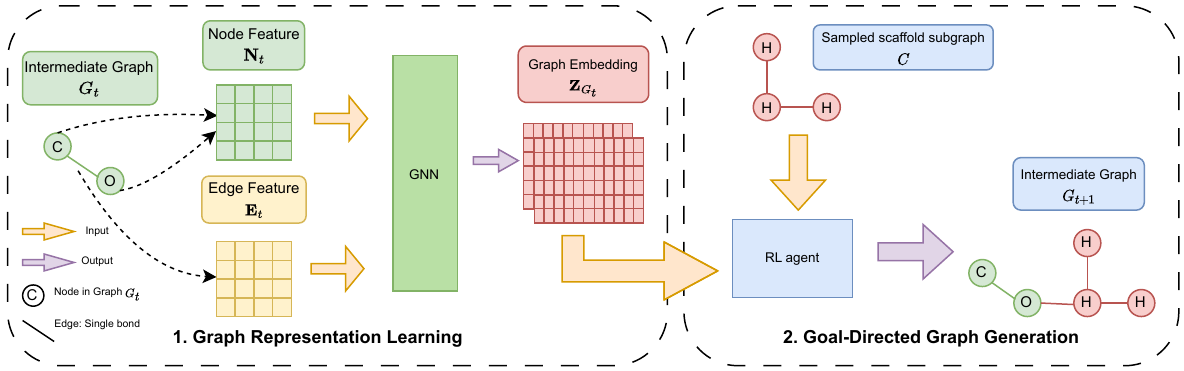}
\caption{An overview of the GCPN model: this is an example of iterative graph generation from an intermediate graph $G_{t}$ to an intermediate graph $G_{t+1}$. Part 1 is the illustration of the graph representation learning process based on a GNN. Part 2 is the illustration of the graph generative procedure based on a reinforcement learning (RL) agent. New nodes or edges are marked in red.}
\label{gnn1}
\vspace*{-\baselineskip}
\end{figure}

In this section, we provide the theoretical background for graph generative models. A GNN-based graph generative model consists of a GNN model and a graph generative framework. The GNN learns the hidden representations of a graph, such as the node features and the graph features. The goal of the graph generation framework is to generate realistic molecular graph structures based on a given generative objective. The detail of GCPN is presented in Figure \ref{gnn1}, and illustrations of the GraphAF and GraphEBM architectures are displayed in Figures \ref{gnn2} and \ref{gnn3} in Appendix \ref{meth} respectively.

\subsection{Preliminaries}
A graph is defined as $G = \left(\mathcal{V}, \mathcal{E}\right)$, where $\mathcal{V}$ denotes the set of nodes, and $\mathcal{E} \subseteq \mathcal{V} \times \mathcal{V}$ denotes the set of edges. A relational graph can be expressed as $G = \left(\mathcal{V}, \mathcal{E}, \mathcal{R}\right)$ where $\mathcal{R}$ denotes the set of edge relations or edge types. For example, $(i, j, r)$ means the edge from node $i$ to node $j$ with edge type $r$. A molecular graph can be represented by a tuple of features $\left(\textbf{A}, \textbf{H}, \textbf{E}, \textbf{R} \right)$, where

\begin{itemize}
	\item $\textbf{A} \in \mathbb{R}^{|\mathcal{V}| \times |\mathcal{V}|}$ is the adjacency matrix, with each entry $a_{ij}$ representing an edge (if any) between nodes $i$ and $j$; note that this is different from the conventional $\{0,1\}^{|\mathcal{V}| \times |\mathcal{V}|}$ adjacency matrix format, since there are different types of bonds (i.e., single, double, triple, aromatic).
	\item $\textbf{H} \in \mathbb{R}^{|\mathcal{V}| \times d}$ is the feature matrix, $\textbf{h}_{i} \in \mathbb{R}^{d}$ is the $d$-dimensional features of node $i$.
	\item $\textbf{E} \in \mathbb{R}^{|\mathcal{E}| \times d_{e}}$ is the edge feature matrix, $\textbf{e}_{ij} \in \mathbb{R}^{d_{e}}$ is the $d_{e}$-dimensional features of edge $(i, j)$.
    \item $\textbf{R} \in \mathbb{N}^{|\mathcal{E}|}$ is a vector containing the edge types of each edge $(i, j) \in \mathcal{E}$ and $r_{(i,j)} \in \mathcal{R}$. This feature vector is explicitly used in R-GCN~\cite{RGCN} and GearNet~\cite{GearNet} (details in Appendix~\ref{appn}).
\end{itemize}

The degree matrix $\textbf{D} \in \mathbb{R}^{|\mathcal{V}| \times |\mathcal{V}|}$ of a graph $G = (\mathcal{V}, \mathcal{E})$ is a diagonal matrix with each diagonal entry $d_{ii} = \text{deg}(v_{i})$, which counts the number of edges terminating at a node in an undirected graph.

\subsubsection{Subgraphs \& Isomorphism}
A graph $G^{\prime} = (\mathcal{V}^{\prime}, \mathcal{E}^{\prime})$ is a \textit{subgraph} of a graph $G = (\mathcal{V}, \mathcal{E})$ (denoted $G^{\prime} \subseteq G$) if and only if $\mathcal{V}^{\prime} \subseteq \mathcal{V}$ and $\mathcal{E}^{\prime} \subseteq \mathcal{E}$.
Two graphs $G = \left(\mathcal{V}, \mathcal{E}\right)$ and $G^{\prime} = (\mathcal{V}^{\prime}, \mathcal{E}^{\prime})$ are \textit{isomorphic} (denoted $G \cong G^{\prime}$) if and only if there exists an adjacency-preserving bijective mapping $f: \mathcal{V} \rightarrow \mathcal{V}^{\prime}$, i.e., 
\begin{equation}
\forall i, j \in \mathcal{V}.\ (i, j) \in \mathcal{E} \Longleftrightarrow \left(f(i), f(j)\right) \in \mathcal{E}^{\prime}
\label{bijection}
\end{equation}
An \textit{automorphism} of a graph $G = \left(\mathcal{V}, \mathcal{E}\right)$ is an isomorphism that maps $G$ onto itself.

\subsection{Graph Neural Networks}

All the GNNs investigated in this paper can be abstracted as Message Passing Neural Networks (MPNNs). A general MPNN operation iteratively updates the node features $\textbf{h}^{(l)}_{i} \in \mathbb{R}^{d}$ from layer $l$ to layer $l + 1$ via propagating messages through neighbouring nodes $j \in \mathcal{N}_{i}$, which can be formalised as 
\begin{equation}
\textbf{h}_{i}^{(l+1)} = \text{UPDATE}\left(\textbf{h}_{i}^{(l)}, \mathop{\bigoplus}\limits_{j \in \mathcal{N}_{i}}\text{MESSAGE}\left(\textbf{h}_{i}^{(l)}, \textbf{h}_{j}^{(l)}, \textbf{e}_{ij}\right)\right)
\label{1}
\end{equation}
where MESSAGE and UPDATE are learnable functions, such as Multi-Layer Perceptrons (MLPs), $\mathcal{N}_{i} = \{j|(i,j) \in \mathcal{E}\}$ is the (1-hop) neighbourhood of node $i$, and $\bigoplus$ is a permutation-invariant local neighbourhood aggregation function, such as sum, mean or max. After $k$ iterations of aggregation, node $i$’s representation $\textbf{h}_{i}^{(k)}$ can capture the structural information within its $k$-hop graph neighbourhood. Then, the graph embedding $\textbf{h}_{G} \in \mathbb{R}^{d}$ can be obtained via a READOUT function:
\begin{equation}
\textbf{h}_{G} = \text{READOUT}_{i \in \mathcal{V}}\left(\textbf{h}_{i}^{(k)}
\right)
\label{2}
\end{equation}
which aggregates the node features to obtain the entire graph’s representation $\textbf{h}_{G}$.

Ideally, a maximally powerful GNN could distinguish different graph structures by mapping them to different representations in the embedding space. This ability to map any two different graphs to different embeddings, however, implies solving the challenging graph isomorphism problem. That is, we want isomorphic graphs to be mapped to the same representation and non-isomorphic ones to different representations. Thus, the expressiveness of a GNN is defined as the ability to distinguish non-isomorphic graphs, and can be analysed by comparing to the 1-WL graph isomorphism test, which are summarised in Appendix \ref{appn}.

\subsection{Graph Generative Frameworks}
In this paper, GCPN~\cite{GCPN} and GraphAF~\cite{GraphAF} are used as autoregressive generative frameworks, and GraphEBM~\cite{ebm} is used as the one-shot generative framework for molecular graph generation tasks. GCPN and GraphAF formalise the problem of goal-directed graph generation as a sequential decision process through RL, i.e. the decisions are generated from the generation policy network. GraphEBM uses Langevin dynamics~\cite{lan} to train the proposed energy function by approximately maximising likelihood and generate molecular graphs with low energies.

In GCPN, the iterative graph generation process is formulated as a general decision process: $M = (\mathcal{S}, \mathcal{A}, P, R, \gamma)$, where $\mathcal{S} = \{s_{i}\}$ is the set of states that consists of all possible intermediate and final graphs; $\mathcal{A} = \{a_{i}\}$ is the set of actions that describe the modification made to the current graph at each time step determined by a group of MLPs predicting a distribution of actions; $P$ is the Markov transition dynamics that specifies the possible outcomes of carrying out an action, $p(s_{t+1}|s_{t}, \cdots, s_{0}, a_{t}) = p(s_{t+1}|s_{t}, a_{t})$; $R(s_{t})$ is a reward function that specifies the reward after reaching state $s_{t}$; and $\gamma$ is the discount factor. In each iteration, GCPN takes the intermediate graph $G_t$ at time $t$ and the collection of proposed scaffold subgraphs $C$ as inputs, and outputs the action $a_{t}$, which predicts a new link to be added to $G_t$.

GraphAF defines an invertible transformation from a base distribution (e.g. multivariate Gaussian) to a molecular graph structure $G = (\mathcal{V}, \mathcal{E})$ as the generation policy network. Starting from an empty graph $G_{0}$, in each step a new node $v_{i}$ is generated based on the current sub-graph structure $G_{i}$, i.e., $p(v_{i}|G_{i})$ by the policy network. Next, the edges between this new node and existing nodes are sequentially generated according to the current graph structure, i.e., $p(\mathcal{E}_{ij}|G_{i}, v_{i}, \mathcal{E}_{i, 1:j-1}, \mathcal{E}_{1:j-1, i})$. This process is repeated until all the nodes and edges are generated. 

In GraphEBM, in order to generate molecules with a specific desirable property at one time, it parameterises an energy function $E_{\theta}(G)$ by a GNN, assigning lower energies to data points that correspond to desirable molecular graphs and higher energies to other data points. After training the energy function network $E_{\theta}(G)$, it initialises a random sample $G'$ and applies $K$ steps of Langevin dynamics~\cite{lan} to obtain a desirable data point $G''$ that has low energy.

\section{Evaluation}
\label{eva}
\subsection{Experimental Setup}

\paragraph{Dataset} We use the ZINC-250k~\cite{zinc} dataset for both pre-training and fine-tuning proposed models for its pharmaceutical relevance, moderate size, and popularity. All molecules are presented in the kekulised form with hydrogen removed. ZINC-250k contains around 250,000 drug-like molecules with 9 atom types and 3 edge types, and with a maximum graph size of 38 sampled from the ZINC database. The molecules in ZINC are readily synthesisable molecules: it contains over 120 million purchasable ``drug-like'' compounds---effectively all organic molecules that are for sale---a quarter of which are available for immediate delivery. Other datasets, such as QM9~\cite{qm9} (a subset of GDB9~\cite{gdb9}), contain, at least to some degree, virtual molecules that are likely to be synthesisable but have not been made yet, including many molecules with complex annulated ring systems~\cite{dataset1}. In addition, the original works of GCPN, GraphAF, GraphEBM and the benchmark~\cite{benchmark} we used to compare graph generative models have also been trained on ZINC-250k. Thus, ZINC-250k is well-suited for models to learn representations of drug-like and synthesisable molecules in de-novo molecule generation. 

\paragraph{Implementation details} 
\label{imple}
We use the open-source platform TorchDrug~\cite{td} and DIG~\cite{dig} in dataset preparation and graph generative models training. Advanced GNNs are implemented in PyTorch~\cite{pytorch} in the MPNN framework aligned with TorchDrug~\cite{td}. The basic architectures of GCPN and GraphAF are implemented in TorchDrug. Note that the original GraphAF cannot allow GNN module to aggregate edge features in the message-passing mechanism, since the intermediate molecular graph generated by the autoregressive flow doesn't contain edge features. Therefore, we propose an improved version of GraphAF considering edge features, named GraphAF+e. To conduct a fairly analogous evaluation on GCPN, we considered GCPN without considering edge features, named GCPN--e. A detailed description of how to incorporate edge features is in Appendix \ref{appa}. The basic architectures of GraphEBM are implemented in DIG~\cite{dig}.

In the experiments, we replace R-GCN in GCPN (both with and without edge features), GraphAF (both with and without edge features), and GraphEBM with six more expressive GNNs: GIN~\cite{HPGNN}, GAT~\cite{GAT}, GATv2~\cite{GATv2}, PNA~\cite{PNA}, GSN~\cite{GSN} and GearNet~\cite{GearNet}. We set all GNNs in the experiments to have 3 hidden layers with batch normalisation~\cite{bn} and ReLU~\cite{relu} activation applied after each layer. We find little improvement when further adding GCN layers. Each GNN uses a linear layer to transform the edge features in the graph to a hidden embedding, and concatenates with the node features for message passing. In addition, they use sum as the READOUT function for graph representations. For PNA and GSN, the MESSAGE and UPDATE functions are parameterised by linear layers. For GSN, we set the graph substructure set to contain cyclic graphs of sizes between 3 and 8 (both inclusive), which are some of the most important substructures in molecules~\cite{GSN}. For GAT and GATv2, we use multi-head attention with $k=3$ after a manual grid search for attention heads.

The autoregressive generative models (GCPN and GraphAF), with all different GNNs, are pre-trained on the ZINC-250k dataset for 1 epoch, as we do not observe too much performance gain from increasing pre-training epochs. They are then fine-tuned towards the target properties with RL, using the proximal policy optimisation (PPO) algorithm~\cite{ppo}. The models are fine-tuned for 5 epochs for all goal-directed generation tasks with early stopping if all generation results in one batch collapse to singleton molecules. We set the agent update interval for GCPN and GraphAF models to update their RL agents every 3 and 5 batches respectively, as considering the cost of computing resources. During generating process, we allow our RL agents in all models to make 20 resamplings on the intermediate generated graphs if they cannot generate chemically valid molecular graphs. The max graph size is set as 48 empirically. For GCPN, the collection of proposed scaffold subgraphs $C$ for GCPN are sampled from non-overlapping scaffolds in the ZINC-250k dataset. For GraphAF, we define multivariate Gaussian as our base distribution and use node MLPs and edge MLPs which have two fully-connected layers equipped with tanh non-linearity to generate the nodes and edges respectively. We notice limited improvements in performance when increasing the number of MLP layers. Adam~\cite{adam} is used as the optimiser for both pre-training and fine-tuning tasks. Reward temperature for each metric, the number of neurons in each hidden layer and the learning rate for fine-tuning) for each model on each task through grid search by Optuna~\cite{optuna} independently.

In the one-shot generative models (GraphEBM), with all different GNNs, we adopt an energy network of $L = 3$ layers with hidden dimension $d = 64$ for each GNN. For training, we tune the following hyperparameters: the sample step $K$ of Langevin dynamics, the standard deviation of the added noise in Langevin dynamics $\sigma$ and the step size $\lambda$. All models are trained for up to 20 epochs. In addition, we clip the gradient used in Langevin dynamics so that its value magnitude can be less than 0.01. 

The experiments were run with a mix of NVIDIA A100 GPUs with 40GB memory and NVIDIA V100 GPUs with 16GB memory. The total amount of training time for all GCPN, GraphAF and GraphEBM variants under all metrics is around 1650 GPU hours. All details are provided in Appendix~\ref{appa}.

\paragraph{Baselines}
\label{metric}
We compare our proposed models based on the original GCPN (both with and without edge features), GraphAF (both with and without edge features), and GraphEBM on six goal-directed molecule generation tasks: Penalised logP~\cite{plogp}, QED~\cite{qed}, synthetic accessibility (SA)~\cite{sa}, DRD2~\cite{drd2}, Median1 and Median2. All generation metrics are taken from the Therapeutic Data Commons (TDC) \cite{tdc}. Due to the practicality of de-novo molecule design, we only consider the single generation objective for all generation tasks. Specifically, SA stands for how hard or how easy it is to synthesise a given molecule. Penalised logP is a logP score that also accounts for ring size and SA, while QED is an indicator of drug-likeness. DRD2 is derived from a support vector machine (SVM) classifier with a Gaussian kernel fitting experimental data to predict the bioactivities against their corresponding disease targets. Median1~\cite{tdc} measures the average score of the molecule's Tanimoto similarity~\cite{TS} to Camphor and Menthol. Median2~\cite{tdc} measures the average score of the molecule's Tanimoto similarity to Tadalafil and Sildenafil. Penalised logP has an unbounded range, while QED, DRD2, Median1, Median2 and SA have a range of $[0, 1]$ by definition. Higher scores in Penalised logP, QED, DRD2, Median1 and Median2 and a lower score in SA are desired. Note that all scores are calculated from empirical prediction models.

We choose to use DRD2, Median1 and Median2 to evaluate generative models, since they are mature and representative generative metrics~\cite{benchmark, dataset1}. In addition, some other metrics are data-missing or inappropriate and thus cannot reflect the ability of generative models properly. For example, GSK3 cannot evaluate all the generated molecules; multi-property objectives (MPO) measure the geometric means of several scores, which will be 0 if one of the scores is 0; Valsartan Smarts is implemented incorrectly in TDC: it computes the geometric means of several scores instead of arithmetic means of several scores, which lead to incorrect results in the benchmark~\cite{benchmark}.

We compare the best GCPN variant, GraphAF variant, and GraphEBM variant with eight state-of-the-art approaches for molecule generation~\cite{benchmark} on 6 generation metrics: Penalised logP, SA, DRD2, Median1, Median2 and QED. All results of baselines are taken from original papers unless stated.

\subsection{Results}

\subsubsection{Improving GNN-based Graph Generative Methods}

\paragraph{De-novo molecule design with more expressive GNNs}

As we re-evaluate the Penalised logP and QED scores of the top-3 molecules found by GCPN, we note that our results turn out to be higher than the results reported in the original GCPN paper. We hypothesise this to be due to our more extensive hyperparameter searching. In Table \ref{tab:GCPN1} in Appendix~\ref{meth}, we explore a set of simpler generation metrics based on the GCPN framework, such as Penalised logP, QED and SA, which have been widely used as objectives in previous work on GNN-based graph generative models. 

\begin{table}[t]
	\centering
	\caption{Comparison of the top-3 DRD2, Median1 and Median2 scores of the generated molecules by GCPN variants, with the top-3 property scores of molecules in the ZINC dataset for reference.}
	\label{tab:GCPN2}
  \resizebox{\textwidth}{!}{
	\begin{tabular}{lccccccccc}
		\toprule
	\multirow{2.5}{*}{Model} & \multicolumn{3}{c}{DRD2} & \multicolumn{3}{c}{Median1} & \multicolumn{3}{c}{Median2}\\
		\cmidrule(lr){2-4} \cmidrule(lr){5-7} \cmidrule(lr){8-10}
		& \(\mathbf{1st}\) & \(\mathbf{2nd}\) & \(\mathbf{3rd}\) & \(\mathbf{1st}\) & \(\mathbf{2nd}\) & \(\mathbf{3rd}\) & \(\mathbf{1st}\) & \(\mathbf{2nd}\) & \(\mathbf{3rd}\)\\
		\midrule
            ZINC & 0.9872  & 0.9815 & 0.9773 & 0.3243 & 0.3096 & 0.3096 & 0.2913 & 0.2765 & 0.2749 \\
            \midrule
		R-GCN & 0.4790  & 0.4790 & 0.4790 &0.3367&0.3367&0.3242 & 0.1921 & 0.1891 & 0.1891 \\
		GIN & 0.3460  & 0.3094 & 0.3094 &0.3243&0.3243&0.3235& 0.1770 & 0.1766 & 0.1730\\
		GAT  & 0.4946  & 0.4946 & 0.4946 & 0.3367&0.3367&0.3328 & 0.1648 & 0.1640  & 0.1637 \\
            GATv2 & 0.5101  & 0.4946 & 0.4946 & 0.3367&0.3367&0.3331 & 0.1759 & 0.1720 & 0.1697\\
            PNA & 0.5828  & 0.4448 & 0.4448 &0.3472&0.3254&0.3254 & 0.1629 & 0.1619 & 0.1605\\
            GSN & 0.5363  & 0.4946 & 0.4946 &0.3243&0.3243&0.3235& 0.1770 & 0.1766 & 0.1730\\
            GearNet & \color{DarkGreen}\textbf{0.9696}  & \color{DarkGreen}\textbf{0.9684} & \color{DarkGreen}\textbf{0.9404} &\color{DarkGreen}\textbf{0.3367}&\color{DarkGreen}\textbf{0.3367}&\color{DarkGreen}\textbf{0.3367} & \color{DarkGreen}\textbf{0.2862} & \color{DarkGreen}\textbf{0.2794} & \color{DarkGreen}\textbf{0.2794} \\
		\bottomrule
	\end{tabular}}
\end{table}

As summarised in Table \ref{tab:GCPN1}, after replacing the inner R-GCN in GCPN with more expressive GNNs, we can observe a significant improvement of GCPN in Penalised logP: GCPN with GIN, GearNet and GSN can achieve the saturated 11.19 in Penalised logP. Moreover, with GearNet, the performance of the GCPN variants can outperform the original GCPN on all three metrics. However, we find that \emph{these benchmarks are not suited to differentiate between different models}, since we see many GCPN variants with different GNNs achieve similar and saturated results on those metrics. In addition, these metrics are not representative enough to obtain meaningful conclusions, as discussed in Section ~\ref{rel1}. Therefore, \emph{there is a need of better graph generation objectives}.

\paragraph{De-novo molecule design with better graph generation objectives}
From Table \ref{tab:GCPN1}, we notice that both Penalised logP and SA are saturating on advanced GNNs, making them ineffective metrics for distinguishing the capability of different GNN models, and we need better de-novo molecule generation metrics. Therefore, we introduce three more representative metrics: DRD2, Median1 and Median2, as described in Section~\ref{metric}, and report the top-3 property scores of molecules generated by each model trained on those three metrics in Table \ref{tab:GCPN2} (GCPN), Table \ref{tab:GraphAF1} (GraphAF), and Table \ref{tab:GraphEBM} (GraphEBM) in Appendix~\ref{meth}. The results show that GCPN, GraphAF, and GraphEBM with more expressive GNNs, such as GearNet, can outperform the original GCPN, GraphAF and GraphEBM with R-GCN on all generation tasks by a significant margin. Specifically, on metrics such as DRD2 and Meidan2, GCPN, GraphAF, and GraphEBM with more expressive GNNs can vastly improve the original performance. This observation further indicates that by combining with more expressive GNNs, GCPN, GraphAF, and GraphEBM can successfully capture the distribution of desired molecules. Therefore, we suggest that DRD2, Median1, Median2 and QED are better graph generation metrics for differentiating different GNNs.

\begin{table}[t]
	\centering
	\caption{Comparison of the top-1 DRD2, Median1, Median2 and QED scores with the selected non-GNN-based generative models. The full table can be found in Table \ref{tab:generative2} in Appendix \ref{meth}.}
	\label{tab:generative1}
 \resizebox{\textwidth}{!}{
	\begin{tabular}{lllll}
		\toprule
	\multirow{1}{*}{Model} & \multicolumn{1}{l}{DRD2} & \multicolumn{1}{l}{Median1} & \multicolumn{1}{l}{Median2} & \multicolumn{1}{l}{QED}\\
            \midrule
            GCPN (R-GCN) & 0.479 & 0.337 & 0.192 & 0.948\\
            GCPN (GearNet) & 0.970 \textbf{($+$102.51$\%$)} & 0.337 \textbf{($+$0.00$\%$)} & 0.286 \textbf{($+$48.96$\%$)} & 0.948 \textbf{($+$0.00$\%$)}\\
            \midrule
            GraphAF (R-GCN) & 0.928 & 0.281 & 0.143 & 0.946\\
            GraphAF (GearNet) & 0.987 \textbf{($+$6.36$\%$)} & 0.290 \textbf{($+$3.20$\%$)} & 0.183 \textbf{($+$27.97$\%$)} & 0.947 \textbf{($+$0.11$\%$)}\\
            \midrule
            GraphEBM (R-GCN) & 0.691 & 0.281 & 0.148 & 0.948\\
            GraphEBM (GearNet) & 0.944 \textbf{($+$36.61$\%$)} & 0.400 
            \textbf{($+$42.35$\%$)} & 0.207 \textbf{($+$39.86$\%$)} & 0.948 \textbf{($+$0.00$\%$)}\\
            \midrule
            LSTM HC (SMILES)~\cite{dataset1}& 0.999 & 0.388 & 0.339 & 0.948 \\
            DoG-Gen~\cite{DoG-Gen} & 0.999 & 0.322 & 0.297 & 0.948 \\
            GP BO~\cite{GPBO} & 0.999 & 0.345 & 0.337 & 0.947 \\
            SynNet~\cite{SynNet} & 0.999 & 0.244 & 0.259 & 0.948 \\
            GA+D~\cite{GA+D} & 0.836 & 0.219 & 0.161 & 0.945 \\
            VAE BO (SMILES)~\cite{SMILES-VAE} & 0.940 & 0.231 & 0.206 & 0.947 \\
            Graph MCTS~\cite{Graph-MCTS} & 0.586 & 0.242 & 0.148 & 0.928 \\
            MolDQN~\cite{MolDQN} & 0.049 & 0.188 & 0.108 & 0.871 \\
		\bottomrule
	\end{tabular}}
\end{table}

\paragraph{Comparison with non-GNN-based graph generative methods}
We report the top-1 DRD2, Median1, Median2 and QED scores found by all the GNN-based and non-GNN-based graph generative models in Table \ref{tab:generative1}. As displayed in Table \ref{tab:generative1}, original GCPN, GraphAF, and GraphEBM are not competitive among graph generative models on all the goal-directed molecule generation tasks. However, after modifying their inner GNNs with more advanced GNNs, such as GearNet, they can outperform or match state-of-the-art results across other generative approaches on the de-novo molecule generation task. Specifically, replacing R-GCN with GearNet can achieve an average of 50.49\% improvement on GCPN, 12.51\% on GraphAF, and 39.61\% on GraphEBM in de-novo molecule design, with the proposed generation metrics. 

Visualisations of the generated molecules with desired generative metrics by GCPN and GraphAF variants are presented in Figure \ref{c} in Appendix \ref{appb}. In addition, we only report eight selected graph generative methods in the benchmark~\cite{benchmark} in Table \ref{tab:generative1}, and GCPN with GearNet can achieve comparable results across 17 other non-GNN-based graph generative methods on the proposed metrics, which are fully reported in Table \ref{tab:generative2} in Appendix \ref{meth}. 

\subsubsection{Correlation Between GNN Expressiveness and Graph Generation}
\label{corre}

\begin{table}[t]
	\caption{Graph classification metrics among GCPN and GraphAF variants. $\mathit{NLL}_\mathit{all}$ means the average negative log likelihood loss for all node and edge classification tasks. $\mathit{Acc}$ means the average accuracy for all node and edge classification tasks. $\mathit{NLL}_{e}$ and $\mathit{NLL}_{n}$ mean the average negative log likelihood evaluated on classification tasks for edge and node respectively. By considering $\mathit{Acc}$, we derive a ranking of all GNNs in our practical setting in decreasing order of GNN expressiveness: GSN, PNA, GIN, GearNet, R-GCN, GATv2 and GAT. By considering $\mathit{NLL}_{e}$, we derive another ranking of the GNNs in decreasing order of one-hot encoding edge feature extraction ability: GearNet, R-GCN, PNA, GSN, GIN, GATv2, and GAT (More details in Appendix~\ref{appn}).}
    \vspace*{0.5\baselineskip}
    \centering
	\label{tab:acc1}
	\begin{tabular}{lcccc}
		\toprule
        \multirow{2}{*}{Model} & \multicolumn{2}{c}{GCPN} & \multicolumn{2}{c}{GraphAF} \\
		\cmidrule(lr){2-3} \cmidrule(lr){4-5} 
		& $\mathit{NLL}_\mathit{all}$ & $\mathit{Acc}$ & $\mathit{NLL}_{e}$ & $\mathit{NLL}_{n}$ \\
        \midrule
        GearNet & 2.1265 $\pm$ 0.0017 & 0.8677 $\pm$ 0.0003 & {\color{DarkGreen}\textbf{0.9275 $\pm$ 0.0048}} & {\color{DarkGreen}\textbf{2.8981 $\pm$ 0.0398}}\\
        R-GCN & 2.1427 $\pm$ 0.1326 & 0.8656 $\pm$ 0.0088& 1.0674 $\pm$ 0.0204 & 3.3312 $\pm$ 0.1557\\
        PNA & 2.1172 $\pm$ 0.0036 & 0.8716 $\pm$ 0.0003& 1.1043 $\pm$ 0.1148 & 3.2840 $\pm$ 0.2154 \\
        GSN & {\color{DarkGreen}\textbf{1.4219 $\pm$ 0.0026}} & {\color{DarkGreen}\textbf{0.9308 $\pm$ 0.0002}}& 1.1319 $\pm$ 0.0678 & 3.1280 $\pm$ 0.0566\\  
        GIN & 2.1500 $\pm$ 0.3158 & 0.8680 $\pm$ 0.0220& 1.1691 $\pm$ 0.1167 & 3.6067 $\pm$ 0.2589\\
        GATv2 & 2.8285 $\pm$ 0.2878 & 0.8013 $\pm$ 0.0244& 1.2426 $\pm$ 0.0370 & 3.1325 $\pm$ 0.0681 \\
        GAT & 2.8705 $\pm$ 0.2354 & 0.7957 $\pm$ 0.0211& 1.2737 $\pm$ 0.0340 & 3.1541 $\pm$ 0.0844\\
		\bottomrule
	\end{tabular}
    \vspace*{0.5\baselineskip}
\end{table}

In the pre-training phase, the GNNs are used to predict all node types and edge types in all masked graphs in the training data. We report the graph classification results for all GNNs in Table \ref{tab:acc1}. In the GCPN framework, we can see GSN perform the best with 93.08\% accuracy on the graph classification task. In the GraphAF framework, we can see GearNet perform the best with the lowest edge and node average negative log likelihood: 0.9275 and 2.8981 respectively. Both GSN and GearNet are more expressive GNNs than R-GCN demonstrated in Appendix \ref{appn}. It is not surprising that expressive GNNs can outperform other GNNs on the graph classification task, since the expressiveness of a GNN is defined as the ability to distinguish non-isomorphic graphs.

However, by comparing Table \ref{tab:acc1} with Tables \ref{tab:GCPN3}, \ref{tab:GCPN2}, \ref{tab:GraphAF1}, \ref{tab:GraphAF2}, it is worth noting that \emph{more expressive GNNs cannot ensure better performance of GNN-based graph generative models in molecular generation tasks}. For example, PNA and GSN perform better than R-GCN on graph classification tasks (Table~\ref{tab:acc1}), but GCPN with PNA or GSN cannot surpass the original GCPN with R-GCN on all generation metrics (Table~\ref{tab:GCPN2}). As also demonstrated in Appendix \ref{appn}, we find graph generation models with strong one-hot encoding edge feature extraction ability GNNs, such as GearNet and R-GCN, can perform better than models with other GNNs, such as GATv2 and GAT. Details about GNN expressiveness and edge feature extraction ability are described in Appendix \ref{appn}. We conclude that the graph generation tasks requires other abilities of GNNs than graph prediction tasks, such as edge feature extraction.

\paragraph{De-novo molecule design with GNNs incorporating edge features}
We investigate the performance of GCPN and GraphAF using GNNs with and without edge features. It is worth mentioning that the original GCPN considers edge features but GraphAF does not. 
The results of the GCPN without edge features (GCPN--e) are summarised in Table \ref{tab:GCPN3} in Appendix \ref{meth}, and those of the original GCPN are in Table \ref{tab:GCPN2}. Both sets of results demonstrate that including edge features can significantly improve the top-3 scores on all three metrics for most GNNs. For instance, with the help of edge feature extraction, GCPN with GIN, GAT, GATv2, PNA and GSN can improve GCPN--e by 24.0\%, 149.8\%, 70.5\%, 31.0\% and 12.0\% respectively on the top-1 score on the metric DRD2. 

The original GraphAF comes without edge feature extraction and we improved it by incorporating the edge features. Table \ref{tab:GraphAF1} (the orginal GraphAF) and Table \ref{tab:GraphAF2} (GraphAF+e) in Appendix \ref{meth} summarised the results of both implementations accordingly. The results align with the GCPN case, where GraphAF+e improves the top-3 scores on all three metrics for most GNNs than the original GraphAF. With the help of edge feature extraction, GraphAF+e with GAT, GATv2, PNA and GSN can increase by 60.0\%, 20.0\%, 230.4\% and 312.4\% respectively on the top-1 score on the metric DRD2, compared with the original GraphAF with those GNNs.

The results above indicate the importance of \emph{aggregating edge features for GNN-based generative models}. By harnessing the power to extract knowledge from edge information, the potential for generating more refined and accurate graphs is enhanced. We also notice that, when the GNNs used are either R-GCN or GearNet, we find GCPN and GraphAF perform well with and without considering edge information on generation metrics except DRD2, so we hypothesise the reason is that they absorb the number of edge relations as prior information in the model. 

In summary, we conclude the following results:

\begin{enumerate}
    
    \item More expressive GNN can lead to better results in the graph classification task. However, GNN expressiveness is not a necessary condition for a good GNN-based generative model. Generation tasks require other abilities of GNNs, such as edge feature extraction.

    \item Although Penalised logP and QED are widely used for generative metrics in evaluating goal-directed graph generative models, they are not effective metrics to differentiate different generative models. Other metrics, such as DRD2, Median1 and Median2, can better evaluate the ability of a graph generative model. Under those metrics, we can see the performance of GCPN, GraphAF and GraphEBM can be enhanced by using more robust GNNs.

    \item After applying advanced GNN to current GNN-based graph generative methods, such as GCPN, GraphAF and GraphEBM, they can outperform or match state-of-the-art results across 17 other generative approaches in the de-novo molecule generation task.

\end{enumerate}

\section{Limitation and Conclusion}
Due to computation cost, we acknowledge several limitations of the current study: we cannot exhaustively explore every method, such as other Relational GNNs, and thoroughly tune every hyperparameter; we cannot evaluate all generative models on other complicated datasets besides ZINC-250k, such as ChEMBL~\cite{chembl}, and other generation metrics~\cite{benchmark}. However, our efforts have still provided valuable insights into investigating the expressiveness of GNN on the graph generation task, because of our focus on many different generative models and diverse generation objectives.

After exploring (1) unexplored underlying GNNs and (2) non-trivial generative objectives, we would like to conclude that expressiveness is not a necessary condition for a good GNN-based generative model. By evaluating GCPN variants, GraphAF variants, and GraphEBM variants on effective metrics, we demonstrate that GNN-based generative methods (GCPN, GraphAF and GraphEBM) can be improved by using more robust GNNs (e.g. strong edge features extraction and edge relation detection). With more robust GNNs, GNN-based graph generative models can outperform or match state-of-the-art results across 17 other generative approaches on de-novo molecule design tasks~\cite{benchmark}. In the future, we plan to explore the necessary conditions for GNN to enhance the performance of GNN-based graph generative models.
\section*{Acknowledgement}
This work was performed using the Sulis Tier 2 HPC platform hosted by the Scientific Computing Research Technology Platform at the University of Warwick, and the Cirrus UK National Tier-2 HPC at EPCC. Sulis is funded by EPSRC Grant EP/T022108/1 and the HPC Midlands+ consortium. Cirrus is funded by the University of Edinburgh and EPSRC Grant EP/P020267/1. For the purpose of open access, the authors have applied a Creative Commons Attribution (CC BY) licence to any Author Accepted Manuscript version arising.

\bibliographystyle{unsrtnat}
\bibliography{reference}

\clearpage
\appendix
\section{Appendix: Non-GNN-based Graph Generative Models}
\label{ngnn}
\paragraph{Genetic Algorithm (GA)} 
The Genetic Algorithm (GA) is a widely used heuristic technique that draws inspiration from natural evolutionary mechanisms. It combines mutation and/or crossover perturbing a mating pool to enable exploration in the design space. SynNet~\cite{SynNet} utilises a genetic algorithm on binary fingerprints and subsequently decodes them into synthetic pathways. GA+D~\cite{GA+D} constitutes a genetic algorithm improved through the incorporation of a neural network (DNN) based discriminator model.

\paragraph{Bayesian Optimisation (BO)~\cite{BO}}
Bayesian Optimisation (BO) represents a broad category of techniques that constructs a surrogate model for the objective function through the application of Bayesian machine learning methods like Gaussian process (GP) regression. It then employs an acquisition function that integrates information from the surrogate model and its associated uncertainty to determine optimal sampling points. GPBO~\cite{GPBO} optimises the GP acquisition function with Graph GA methods in an inner loop. 

\paragraph{Variational Autoencoders (VAEs)~\cite{VAE}}
Variational Autoencoders (VAEs) belong to a category of generative techniques that focus on maximising a lower bound of the likelihood, known as the evidence lower bound (ELBO), as opposed to directly estimating the likelihood itself. A VAE typically learns to map molecules to and from real space to enable the indirect optimisation of molecules by numerically optimising latent vectors.
SMILES-VAE~\cite{SMILES-VAE} uses a VAE to model molecules represented as SMILES strings. 

\paragraph{Monte-Carlo Tree Search (MCTS)}
Monte-Carlo Tree Search (MCTS) conducts a localised and stochastic exploration of each branch originating from the present state, which could be a molecule or an incomplete molecule. It then identifies the most promising branches, which usually exhibit the highest property scores, to be considered for the subsequent iteration. Graph-MCTS~\cite{Graph-MCTS} is an MCTS algorithm based on atom-level searching over molecular graphs.

\paragraph{Hill Climbing (HC)}
Hill Climbing (HC) is an iterative learning method that incorporates the generated high-scored molecules into the training data and fine-tunes the generative model for each iteration. SMILES-LSTM~\cite{dataset1} leverages a LSTM to learn the molecular distribution represented in SMILES strings, and modifies it to a SELFIES version. DoG-Gen~\cite{DoG-Gen} instead learn the distribution of synthetic pathways as Directed Acyclic Graph (DAGs) with an RNN generator.

\paragraph{Reinforcement Learning (RL)}
In molecular design, a state is usually a partially generated molecule; actions are manipulations at the level of graph or string representations; rewards are determined by the desired properties of the molecules generated. MolDQN~\cite{MolDQN} uses a deep Q-network to generate molecular graphs in an atom-wise manner.

\clearpage
\section{Appendix: Details of Generation Metrics} 
The details about the de-novo molecular generation metrics we use~\cite{tdc, benchmark} are described as follows:
\begin{enumerate}
    \item \textbf{Penalised LogP (PlogP):} It is a logP score that also accounts for ring size and synthetic accessibility (SA). The penalised logP score measures the solubility and synthetic accessibility of a compound~\cite{plogp}. It can range between zero (all properties unfavourable) and an unbounded limit (all properties favourable).
    \item \textbf{Quantitative Estimate of Drug-likeness (QED):} The empirical rationale of the QED measure reflects the underlying distribution of molecular properties including molecular weight, logP, topological polar surface area, number of hydrogen bond donors and acceptors, the number of aromatic rings and rotatable bonds, and the presence of unwanted chemical functionalities~\cite{qed}. It can range between zero (all properties unfavourable) and one (all properties favourable).
    \item \textbf{Synthetic Accessibility (SA):} Synthetic Accessibility score stands for how hard or how easy it is to synthesise a given molecule, based on a combination of the molecule’s fragments contributions. The oracle is caluated via RDKit, using a set of chemical rules~\cite{sa}. The method is based on the combination of molecule complexity and fragment contributions obtained by analyzing structures of a million already synthesised chemicals, and in this way captures also historical synthetic knowledge. It can range between zero (all properties favourable) and one (all properties unfavourable).
    \item \textbf{DRD2:} it uses dopamine type 2 receptor as the biological target. The oracle is constructed by using a support vector machine classifier with a Gaussian kernel and ECFP6 fingerprints on the ExCAPE-DB dataset to measure a molecule's biological activity against dopamine~type~2 receptor \cite{drd2}. It ranges between zero (all properties unfavourable) and one (all properties~favourable).
    \item \textbf{Median1:} it measures the average score of the molecule's Tanimoto similarity to Camphor and Menthol~\cite{tdc}. In the median molecules benchmarks, the similarity to several molecules has to be maximised simultaneously. Besides measuring the obtained top score, it is instructive to study if the models also explore the chemical space between the target structures~\cite{dataset1}. It can range between zero (all properties unfavourable) and one (all properties favourable).
    \item \textbf{Median2:} it measures the average score of the molecule's Tanimoto similarity to Tadalafil and sildenafil~\cite{tdc}. In the median molecules benchmarks, the similarity to several molecules has to be maximised simultaneously. Besides measuring the obtained top score, it is instructive to study if the models also explore the chemical space between the target structures~\cite{dataset1}. It can range between zero (all properties unfavourable) and one (all properties favourable).
\end{enumerate}
\clearpage
\section{Appendix: Graph Neural Networks}
\label{appn}

\subsection{GNN Architectures}

\paragraph{Relational Graph Convolutional Network (R-GCN)}
The graph convolution operation of the original GCN ~\cite{GCN} can be defined as follows:

\begin{equation}
\textbf{h}_{i}^{(l+1)} = \sigma\left(\sum_{j \in \mathcal{N}_{i}}c_{ij}\textbf{W}^{(l)}\textbf{h}_{j}^{(l)}\right)
\label{eGCN}
\end{equation}

where $c_{ij}$ is a normalisation constant for each edge $\mathcal{E}_{ij}$ which originates from using the symmetrically normalised adjacency matrix $\textbf{D}^{-\frac{1}{2}}\textbf{A}\textbf{D}^{-\frac{1}{2}}$, $\textbf{W}^{(l)}$ is a learnable weight matrix, and $\sigma$ is a non-linear activation function. R-GCN ~\cite{RGCN} makes use of the relational data of the graphs, and extends the graph convolution operation to the following: let $\mathcal{R}$ be the edge relation type (for molecular graphs, this can be the bond type), then

\begin{equation}
\textbf{h}_{i}^{(l+1)} = \sigma\left(\sum_{r \in \mathcal{R}}\sum_{j \in \mathcal{N}^{r}_{i}}c_{i,r}\textbf{W}_{r}^{(l)}\textbf{h}_{j}^{(l)}+\textbf{W}_{0}^{(l)}\textbf{h}_{i}^{(l)}\right)
\label{eRGCN}
\end{equation}

where $\mathcal{N}^{r}_{i}$ denotes the set of neighbouring nodes of node $i$ under relation $r \in \mathcal{R}$, $c_{i,r}$ is a problem-specific normalisation constant that can either be learnt or chosen in advance and $\textbf{W}_{r}^{(l)}$ denotes the learnable matrix for edge type $r$. It has been shown that \textit{neither GCN nor R-GCN is as expressive as the 1-WL test}~\cite{HPGNN}.

\paragraph{Graph Isomorphism Network (GIN)}
Each GIN ~\cite{HPGNN} layer updates the node features as follows:

\begin{equation}
\textbf{h}_{i}^{(l+1)} = \phi^{(l)}\left(\left(1+\epsilon^{(l)}\right)\textbf{h}_{i}^{(l)}+\sum_{j \in \mathcal{N}_{i}}\textbf{h}_{j}^{(l)}\right)
\label{eRIN}
\end{equation}

where $\phi^{(l)}$ is an MLP, and $\epsilon^{(l)}$ is a learnable scalar. \textit{GIN is provably as expressive as the 1-WL test},
which makes it one of the maximally-expressive GNNs (proof in~\cite{HPGNN}).

\paragraph{Principal Neighbourhood Aggregation (PNA)}
The PNA~\cite{PNA} operator defines its aggregation function $\bigoplus$ as a combination of neighbourhood-aggregators and degree-scalers, as defined by the following equation, with $\otimes$ being the tensor product:

\begin{equation}
\bigoplus = \underbrace{\left[\begin{array}{c}
   \text{identity} \\
   \text{amplification} \\
   \text{attenuation}
\end{array}\right]}_\text{scalers} \otimes \underbrace{\left[\begin{array}{c}
   \text{mean} \\
   \text{max} \\
   \text{min} \\
   \text{std}
\end{array}\right]}_\text{aggregators}
\label{ePNA1}
\end{equation}

The PNA operator can then be inserted into the standard MPNN framework, obtaining the PNA layer:

\begin{equation}
\textbf{h}_{i}^{(l+1)} = \phi^{(l)}\left(\textbf{h}_{i}^{(l)}, \mathop{\bigoplus}\limits_{j \in \mathcal{N}_{i}}\psi^{(l)}\left( \textbf{h}_{i}^{(l)}, \textbf{h}_{j}^{(l)}, \textbf{e}_{ij}\right)\right)
\label{ePNA2}
\end{equation}

where $\phi^{(l)}$ and $\psi^{(l)}$ are MLPs. According to the theorem that \textit{in order to discriminate between multisets of size $n$ whose underlying set is $R$, at least $n$ aggregators are needed} (proof in~\cite{PNA}), PNA pushes its expressivity closer towards the 1-WL limit than GIN, by including more aggregators, thereby increasing the probability that at least one of the aggregators can distinguish different graphs.

\paragraph{Graph Substructure Network (GSN)}
GSN~\cite{GSN} adopts a feature-augmented message passing style by counting the appearance of certain graph substructures and encoding them into the features. The feature augmentation of GSN then works as follows: let $\mathcal{G} = \left\{G_{1}, \cdots, G_{K} \right\}$ be a set of pre-computed small (connected) graphs. For each $G_{k} = (\mathcal{V}_{k}, \mathcal{E}_{k})$ in $\mathcal{G}$, we first find its isomorphic subgraphs $G_{k}^{\prime} = (\mathcal{V}_{k}^{\prime}, \mathcal{E}_{k}^{\prime})$ in $G = (\mathcal{V}, \mathcal{E})$. Then, for each node $i \in V$ and $1 \leq k \leq K$, we count the number of subgraphs $G_{k}^{\prime}$ node $i$ belongs to, as defined by the equation below:

\begin{equation}
x^{\mathcal{V}}_{G_{k}}(i) = \left|\{G_{k}^{\prime} \cong G_{k}|i \in \mathcal{V}_{k}^{\prime} \}\right|
\label{eGSN1}
\end{equation}

We then obtain the node structural features for each node $i \in \mathcal{V}$: $\textbf{x}^{\mathcal{V}}_{i} = \left[x^{\mathcal{V}}_{G_{1}}(i), \cdots,x^{\mathcal{V}}_{G_{k}}(i) \right]\in\mathbb{N}^{K}$. Similarly, we can derive the edge structural features for each edge $(i, j) \in \mathcal{E}$: $\textbf{x}^{\mathcal{E}}_{ij} = \left[x^{\mathcal{E}}_{G_{1}}(i, j), \cdots,x^{\mathcal{E}}_{G_{k}}(i,j) \right] \in \mathbb{N}^{K}$ by counting the numbers of subgraphs it belongs to: 

\begin{equation}
x^{\mathcal{E}}_{G_{k}}(i, j) = \left|\{G_{k}^{\prime} \cong G_{k}|i \in \mathcal{E}_{k}^{\prime} \}\right|
\label{eGSN2}
\end{equation}

The augmented features can then be inserted into the messages and follow the standard MPNN, obtaining two variants of GSN layer, GSN-v (vertex-count) and GSN-e (edge-count):

\begin{equation}
\begin{split}
\textbf{h}_{i}^{(l+1)} = \phi^{(l)}\left(\textbf{h}_{i}^{(l)}, \mathop{\bigoplus}\limits_{j \in \mathcal{N}_{i}}\psi^{(l)}\left( \textbf{h}_{i}^{(l)}, \textbf{h}_{j}^{(l)}, \textbf{x}^{\mathcal{V}}_{i}, \textbf{x}^{\mathcal{V}}_{j}, \textbf{e}_{ij}\right)\right) (\text{GSN-v})\\
\textbf{h}_{i}^{(l+1)} = \phi^{(l)}\left(\textbf{h}_{i}^{(l)}, \mathop{\bigoplus}\limits_{j \in \mathcal{N}_{i}}\psi^{(l)}\left( \textbf{h}_{i}^{(l)}, \textbf{h}_{j}^{(l)}, \textbf{x}^{\mathcal{E}}_{ij}, \textbf{e}_{ij}\right)\right) (\text{GSN-e})
\end{split}
\label{eGSN3}
\end{equation}

where $\phi^{(l)}$ and $\psi^{(l)}$ are MLPs and $\bigoplus$ is a permutation-invariant local neighbourhood aggregation function, such as sum, mean or max. It can be proved that \textit{GSN is strictly more expressive than the 1-WL test, when $G_{k}$ is any graph except for the star graphs (i.e., one center nodes connected to one of multiple outer nodes) of any size, and structural features are inferred by subgraph matching} (proof in~\cite{GSN}). This essentially suggests that GSN is more expressive than R-GCN which is at most as expressive as the 1-WL test in general.

\paragraph{Geometry Aware Relational Graph Neural Network (GearNet)}
GearNet~\cite{GearNet} uses an R-GCN~\cite{RGCN} as a fundamental framework to develop a node features and edge features message passing mechanism. Each GearNet layer updates the node features as follows:

\begin{equation}
\textbf{h}_{i}^{(l+1)} = \sigma\left(\text{BN}\left(\sum_{r \in \mathcal{R}}\textbf{W}_{r}^{(l)}\sum_{j \in \mathcal{N}^{r}_{i}}\textbf{h}_{j}^{(l)}\right)\right)+\textbf{h}_{i}^{(l)} (\text{GearNet-v})
\label{eGea1}
\end{equation}

where $\mathcal{R}$ is the edge relation type, $\mathcal{N}^{r}_{i}$ denotes the set of neighbouring nodes of node $i$ under relation $r \in \mathcal{R}$, BN denotes a batch normalisation layer, $\textbf{W}_{r}^{(l)}$ is the learnable convolutional kernel matrix for edge type $r$, and $\sigma$ is a non-linear activation function.

To model the interactions between edges, we first construct a relational graph $G' = \left(\mathcal{V}', \mathcal{E}', \mathcal{R}'\right)$ among edges. Each node in the graph $G'$ corresponds to an edge in the original graph. $G'$ links edge $(i, j, r_{1})$ in the original graph to edge $(w, k, r_{2})$ if and only if $j = w$ and $i \neq k$. The type of this edge is determined by the angle between $(i, j, r_{1})$ and $(w, k, r_{2})$. \textit{The angular information reflects the relative position between two edges that determines the strength of their interaction}~\cite{GearNet}. Similar to R-GCN, the GearNet edge message passing layer is defined as: 

\begin{equation}
\textbf{e}_{i, j, r_{1}}^{(l+1)} = \sigma\left(\text{BN}\left(\sum_{r \in \mathcal{R'}}\textbf{W}_{r}^{\prime(l)}\sum_{(w, k, r_{2}) \in \mathcal{N'}^{r}_{(i, j, r_{1})}}\textbf{e}_{(i,j,r_{1})}^{(l)}\right)\right) (\text{GearNet-e})
\label{eGea2}
\end{equation}

Similar as Eq.~\eqref{eGea1}, the message function for edge $(i,j,r_{1})$ will be updated by aggregating features from its neighbours $\mathcal{N'}^{r}_{(i, j, r_{1})}$, where $\mathcal{N'}^{r}_{(i, j, r_{1})} = \{(w,k,r_{2}) \in \mathcal{V}'|((w,k,r_{2}), (i, j, r_{1}), r) \in \mathcal{E}'\}$.

Finally, the entire GearNet message passing layer can be expressed as:

\begin{equation}
\textbf{h}_{i}^{(l+1)} = \sigma\left(\text{BN}\left(\sum_{r \in \mathcal{R}}\textbf{W}_{r}^{(l)}\sum_{j \in \mathcal{N}^{r}_{i}}\left(\textbf{h}_{j}^{(l)} + \text{FC}\left(\textbf{e}_{j, i, r_{1}}^{(l)}\right)\right)\right)\right)+\textbf{h}_{i}^{(l)}
\label{eGea3}
\end{equation}

where FC denotes a linear transformation on the message function. GearNet is more expressive than R-GCN for its sparse edge message passing mechanism which encodes spatial information of a graph.

\paragraph{Graph Attention Network (GAT)}
In order to generalise the standard averaging or max-pooling aggregators in GNNs, GAT~\cite{GAT} applies attention-based neighbourhood aggregation as its aggregation function to obtain sufficient expressive power to transform the input features into higher-level features. The normalised masked attention coefficient for node $i$ is defined as:

\begin{equation}
\forall j \in \mathcal{N}_{i}, 
\alpha_{ij} = \frac{\exp\left(\text{LeakyReLU}\left(\textbf{a}\left[\textbf{W}\textbf{h}_{i}\Vert\textbf{W}\textbf{h}_{j}\right]\right)\right)}{\sum_{k\in\mathcal{N}_{i}} \exp\left(\text{LeakyReLU}\left(\textbf{a}[\textbf{W}\textbf{h}_{i}\Vert\textbf{W}\textbf{h}_{k}]\right)\right)}
\label{eGAT1}
\end{equation}

where $\textbf{a}$ is a learnable weight vector, representing the attention mechanism $a$: a single-layer feedforward neural network, $\textbf{W}$ is a learnable input linear transformation’s weight matrix and $\Vert$ represents concatenation operation.

Each $K$ multi-head attention GAT layer updates the node features as follows:

\begin{equation}
\textbf{h}_{i}^{(l+1)} = \mathop{\Vert}\limits_{k=1}^{K}\sigma\left(\sum_{j \in \mathcal{N}_{i}}\alpha_{ij}^{k}\textbf{W}^{k}\textbf{h}_{j}^{(l)}\right)
\label{eGAT2}
\end{equation}

where $\Vert$ represents concatenation, $\alpha_{ij}^{k}$ are normalised attention coefficients computed by the $k$-th attention mechanism $(a^{k})$, and $\textbf{W}^{k}$ is the corresponding input linear transformation’s weight matrix. GAT computes a representation for every node as a weighted average of its neighbours through the attention mechanism, which is more flexible than the neighbourhood aggregation in R-GCN.

\paragraph{Graph Attention Network v2 (GATv2)}
GATv2~\cite{GATv2} adopts a strictly more expressive \textit{dynamic graph attention mechanism}~\cite{GATv2} in its aggregation function to learn the graph representation. The graph attention variant that has a universal approximator attention function.

The normalised masked dynamic attention coefficient for node $i$ is defined as:

\begin{equation}
\forall j \in \mathcal{N}_{i}, 
\alpha_{ij} = \frac{\exp\left(\textbf{a}\left(\text{LeakyReLU}\left(\textbf{W}\cdot\left[\textbf{h}_{i}\Vert\textbf{h}_{j}\right]\right)\right)\right)}{\sum_{k\in\mathcal{N}_{i}} \exp\left(\textbf{a}\left(\text{LeakyReLU}\left(\textbf{W}\cdot[\textbf{h}_{i}\Vert\textbf{h}_{k}]\right)\right)\right)}
\label{eGAT21}
\end{equation}

where $\textbf{a}$ is a learnable weight vector, representing the attention mechanism $a$: a single-layer feedforward neural network, $\textbf{W}$ is a learnable input linear transformation’s weight matrix and $\Vert$ represents concatenation operation.

Each $K$ multi-head attention GATv2 layer updates the node features as follows:

\begin{equation}
\textbf{h}_{i}^{(l+1)} = \mathop{\Vert}\limits_{k=1}^{K}\sigma\left(\sum_{j \in \mathcal{N}_{i}}\alpha_{ij}^{k}\textbf{W}^{k}\textbf{h}_{j}^{(l)}\right)
\label{eGAT22}
\end{equation}

where $\Vert$ represents concatenation, $\alpha_{ij}^{k}$ are normalised dynamic attention coefficients computed by the $k$-th attention mechanism $(a^{k})$, and $\textbf{W}^{k}$ is the corresponding input linear transformation’s weight matrix. It has been proved that \textit{GATv2 – a graph attention variant that has a universal approximator attention function, and is thus strictly more expressive than GAT} (proof in~\cite{GATv2}).

\subsection{GNN Expressiveness}

\paragraph{Weisfeiler-Lehman (1-WL) Graph Isomorphism Test}
Similar to GNNs, the 1-WL test iteratively updates the node embeddings of a graph by neighbourhood
aggregation: for each node $i \in \mathcal{V}$ in a graph, an initial colour $c_{i}^{(0)}$ is assigned, and is iteratively updated using random hashes of sums: 

\begin{equation}
c_{i}^{(t+1)} = \text{HASH}\left(\sum_{j \in \mathcal{N}_{i}}c_{j}^{(t)}\right)
\label{3}
\end{equation}

The 1-WL test terminates when stable node colouring of the graph is reached, and outputs a histogram of colours. Two graphs with different colour histograms are non-isomorphic, and two graphs with the same colour histograms are possibly, but not necessarily, isomorphic. The neighbourhood aggregation in the 1-WL test can also be seen as a form of message passing, with GNNs being the learnable analogue. It has been proved that \textit{message-passing GNNs are at most as expressive as the 1-WL test over discrete features} ~\cite{gn}.

\paragraph{GNN Expressiveness in Graph Isomorphism Tests} We first illustrate a general graph representation learning framework ~\cite{gn} and then formulate GNN expressiveness defined by the 1-WL test in a graph representation learning framework. Finally, we explore the theoretical GNN expressiveness for all GNN models we proposed above and analyse the GNN expressiveness in our practical setting.

\vspace{\parskip}
\begin{definition}[Graph Representation Learning]
\label{grl}
  The feature space is defined as $\mathcal{X} := \mathcal{A} \times \mathcal{B}$, where $\mathcal{A}$ is the space of graph-structured data and $\mathcal{B}$ includes all the node subsets of interest, given a graph, given a graph $G \in \mathcal{A}$. Then, a point in $\mathcal{X}$ can be denoted as $(G, S)$, where $S$ is a subset of nodes that are in $G$. Later, we call $(G, S)$ as a graph representation learning (GRL) example. Each GRL example $(G,S) \in \mathcal{X}$ is associated with a target $y$ in the target space $\mathcal{Y}$. Suppose the ground-truth association function between features and targets is denoted by $f^{*}:\mathcal{X} \rightarrow \mathcal{Y}, f^{*}(G,S)=y$. Given a set of training examples $\phi = \{ ( G^{(i)}, S^{(i)}, y^{(i)})\}^{k}_{i=1}$ and a set of testing examples $\psi = \{ ( \hat{G}^{(i)}, \hat{S}^{(i)}, \hat{y}^{(i)})\}^{k}_{i=1}$, a graph representation learning problem is to learn a GNN model $f$ based on $\phi$ such that $f$ is close to $f^{*}$ on $\psi$.
\end{definition}

Note that many frequently investigated learning problems can be formulated as graph representation learning problems by properly defining $\mathcal{X}$ and $\mathcal{Y}$, such as graph classification and node classification problems ~\cite{gn}. Based on the graph representation learning framework and the 1-WL test, we can define GNN expressiveness~\cite{gn} as below.

\vspace{\parskip}
\begin{definition}[GNN Expressiveness]
\label{exp}
  Consider a feature space $\mathcal{X}$ of a graph representation learning problem and a GNN model $f$ defined on $\mathcal{X}$. Define another space $\mathcal{X}(f)$ as a subspace of the quotient space $X/\cong$ such that for two GRL examples $(G^{(1)}, S^{(1)}), (G^{(2)}, S^{(2)})\in \mathcal{X}(f), f(G^{(1)}, S^{(1)}) \neq f(G^{(2)}, S^{(2)})$. Then, the size of $\mathcal{X}(f)$ characterises the expressiveness of GNN model $f$. For two GNN models $f^{(1)}$ and $f^{(2)}$, if $\mathcal{X}(f^{(1)}) \supset \mathcal{X}(f^{(2)})$, then $f^{(1)}$ is more expressive than $f^{(2)}$.
\end{definition}

Note that the expressive power of GNNs in Definition \ref{exp}, characterised by how a model can distinguish non-isomorphic GRL examples, does not exactly match the traditional expressive power used for feed-forward neural networks in the sense of functional approximation~\cite{gn}. Since the universal approximation theorem~\cite{uni}, current studies have proved that feed-forward neural networks can approximate any function of interest~\cite{appro}. \textit{However, these results have not been applied to GNNs due to the inductive bias imposed by additional constraints on the GNN parameter space}~\cite{gn}.

The expressiveness ability of each GNN model is discussed in a theoretical setting (with a sufficient and optimal number of GNN layers) above~\cite{gn, RGCN, GearNet, GSN, PNA, GATv2}. We can make the following conclusions:
\begin{enumerate}
\item R-GCN is theoretically the least expressive GNN among seven GNN models, since it is not as expressive as the 1-WL test~\cite{RGCN}. Other GNN models in the message-passing mechanism are at most as powerful as the 1-WL test in distinguishing different graph-structured features~\cite{gn}.

\item GearNet is more expressive than R-GCN due to its sparse edge message-passing mechanism which encodes spatial information in a graph~\cite{GearNet}.

\item It has been proved that GATv2 – a graph attention variant that has a universal approximation attention function, and is thus strictly more expressive than GAT (proof in \cite{GATv2}).

\item GIN is provably as expressive as the 1-WL test~\cite{RGCN}. PNA pushes its expressivity closer towards the 1-WL limit than GIN (proof in \cite{PNA}). 

\item It can be proved that GSN is strictly more expressive than the 1-WL test under certain constraints on the input graph space (proof in~\cite{GSN}).
\end{enumerate}

Although there is a gap in comparing the GNN expressiveness among different GNN models based on current theoretical research, we can rank the GNN expressiveness based on our graph classification task. As shown in Table \ref{tab:acc1}, we use the average accuracy for all node and edge classification tasks $\mathit{NLL}_{all}$ to implicitly derive a ranking for all GNN models in our practical setting: GSN, PNA, GIN, GearNet, R-GCN, GATv2 and GAT (in decreasing order of GNN expressiveness), which corresponds to our theoretical analysis above. Thus, we can conclude that GSN, PNA and GearNet generally have stronger expressiveness and GIN, R-GCN, GAT and GATv2 have weaker expressiveness.

\paragraph{GNN Expressiveness in Edge Feature Extraction Ability} GNN expressiveness defined by the 1-WL test in Definition $\ref{exp}$ is not a necessary condition for a good GNN-based generative model, since the GNN expressiveness is defined as the ability to distinguish non-isomorphic graphs without valuing the ability of GNN to distinguish the edge type in graphs ~\cite{gn, RGCN}. Actually, GNN expressiveness in Definition $\ref{exp}$ is weak because distinguishing any non-isomorphic GRL examples does not necessarily indicate that we can approximate any function $f$ defined over $\mathcal{X}$. Thus, in order to better compare different GNNs based on other abilities, we propose the general edge feature extraction ability of GNN models below.

\vspace{\parskip}
\begin{definition}[GNN Edge Feature Extraction Ability]
\label{edgeexp}
 Consider a graph representation learning problem formulated in Definition \ref{grl}. In a GRL example $(G, S)$, $S$ corresponds to a pair of nodes of interest. $G$ for each example can be an induced subgraph around $S$ or the entire graph. The target space $\mathcal{Y}$ can be defined by low-level or high-level features related to edges in $G$. For example, the target space $\mathcal{Y}$ can be defined as the edge type in $S$ or the attention~\cite{GAT} between the edge type and the nodes in $S$. Each GRL example $(G,S) \in \mathcal{X}$ is associated with a target $y$ in the target space $\mathcal{Y}$. Suppose the ground-truth association function between features and targets is denoted by $f^{*}:\mathcal{X} \rightarrow \mathcal{Y}, f^{*}(G,S)=y$. Given a set of training examples $\phi = \{ ( G^{(i)}, S^{(i)}, y^{(i)})\}^{k}_{i=1}$ and a set of testing examples $\psi = \{ ( \hat{G}^{(i)}, \hat{S}^{(i)}, \hat{y}^{(i)})\}^{k}_{i=1}$, a graph representation learning problem is to learn a GNN model $f$ based on $\phi$ such that $f$ is close to $f^{*}$ on $\psi$.
\end{definition}

Based on Definition \ref{edgeexp}, we consider a specific target space $\mathcal{Y}$ and define the GNN one-hot encoding edge feature extraction ability below.

\vspace{\parskip}
\begin{definition}[GNN One-hot Encoding Edge Feature Extraction Ability]
\label{edgeexp2}
 Consider the setting proposed on Definition \ref{edgeexp} and define the target space $\mathcal{Y}$ contains the one-hot encoding of the edge relation between two nodes. Given a set of training examples $\phi = \{ ( G^{(i)}, S^{(i)}, y^{(i)})\}^{k}_{i=1}$ and a set of testing examples $\psi = \{ ( \hat{G}^{(i)}, \hat{S}^{(i)}, \hat{y}^{(i)})\}^{k}_{i=1}$, consider a GNN model $f$ learned based on $\phi$, we define the space $\mathcal{C}(f)$ as a set $\{(\hat{G}^{(i)}, \hat{S}^{(i)})|f(\hat{G}^{(i)}, \hat{S}^{(i)}) = \hat{y}^{(i)}\}^{j}_{i=1}$ consisted of correctly classified test examples. Then, the size of $\mathcal{C}(f)$ characterises the one-hot encoding edge feature extraction ability of the GNN model $f$. For two GNN models $f^{(1)}$ and $f^{(2)}$, if $\mathcal{C}(f^{(1)}) \supset \mathcal{C}(f^{(2)})$, we say that $f^{(1)}$ has stronger ability in edge feature extraction than $f^{(2)}$.
\end{definition}

We can implicitly rank the one-hot encoding edge relation detection ability of GNN models based on the average negative log likelihood evaluated on classification tasks for the edge in Table \ref{tab:acc1} our paper: GearNet, R-GCN, PNA, GSN, GIN, GATv2, GAT (in decreasing order of one-hot encoding edge relation detection ability), which almost corresponds to the performance of their generative models shown in Table \ref{tab:GCPN2} and Table \ref{tab:GraphAF2}. Thus, we conclude that more expressive GNN can lead to better results in the graph classification task. However, GNN expressiveness defined by the 1-WL test is not a necessary condition for a good GNN-based generative model. Generation tasks require other abilities of GNNs, such as edge feature extraction and edge relation detection.

\clearpage
\section{Appendix: Experiment Details}
\label{appa}
\paragraph{Implementation of GNN incorporating edge features}
As illustrated in Equation \ref{1}, a general MPNN can aggregate both node and edge embeddings to update the node embedding of the targeted node. In order to incorporate edge embeddings in GNNs, we use a single-layer MLP $f: \mathbb{R}^{d_{e}} \rightarrow \mathbb{R}^{d}$ to project relevant edge features of the targeted node to the same dimension of node features and then concatenate them to node features during the neighbourhood aggregation step in GNNs. 

In GCPN--e, we use GNNs without incorporating edge features as the representational module in GCPN, i.e. there is no edge feature aggregated during the neighbourhood aggregation step in the message-passing mechanism. The general GNN without considering edge features iteratively updates the node features $\textbf{h}^{(l)}_{i} \in \mathbb{R}^{d}$ from layer $l$ to layer $l + 1$ via propagating messages through neighbouring nodes $j \in \mathcal{N}_{i}$, which can be formalised by the following equation:

\begin{equation}
\textbf{h}_{i}^{(l+1)} = \text{UPDATE}\left(\textbf{h}_{i}^{(l)}, \mathop{\bigoplus}\limits_{j \in \mathcal{N}_{i}}\text{MESSAGE}\left(\textbf{h}_{i}^{(l)}, \textbf{h}_{j}^{(l)}\right)\right)
\label{noedge}
\end{equation}

where MESSAGE and UPDATE are learnable functions, such as Multi-Layer Perceptrons (MLPs), $\mathcal{N}_{i} = \{j|(i,j) \in \mathcal{E}\}$ is the (1-hop) neighbourhood of node $i$, and $\bigoplus$ is a permutation-invariant local neighbourhood aggregation function, such as sum, mean or max.

In the original GraphAF, the intermediate molecular graph generated by the autoregressive flow doesn’t contain edge features. Thus, in GraphAF+e, during the graph generation step, we apply a one-hot encoding to each edge according to the edge type to obtain edge features in the intermediate generated molecular graph. 

\paragraph{Reward design implementation}
For the property optimisation task, we use the same reward design in GraphAF, which incorporates both intermediate and final rewards for training the policy network. A small penalisation will be introduced as the intermediate reward if the edge predictions violate the valency check. The final rewards include both the score of targeted-properties of generated molecules and the chemical validity reward. The final reward is distributed to all intermediate steps with a discounting factor to stabilise the training~\cite{GraphAF}. The property-targeted reward $r$ for a molecule $m$ on a metric $d$ is defined as follows:

\begin{equation}
r(m) = \text{exp}\left(\frac{d(m)}{t}\right)
\label{ereward}
\end{equation}

where $t$ is the temperature for reward design decided by the grid search.

\paragraph{Hyper-parameter tuning}
All the pre-training works are trained with an Adam optimiser with a learning rate of 0.001. For GCPN, we fixed batch sizes for pre-training and fine-tuning as 128 and 32. For GraphAF, we fixed batch sizes for pre-training and fine-tuning as 32 and 32. Each reward scale factor for Penalised logP, QED and SA is set as 1 after a manual grid search in the space \{1, 5, 10\} based on evaluating each generative metric based on the performance of the original GCPN and GraphAF. The reward scale factors for DRD2, Median1 and Median2 are set as 0.5, 0.05 and 0.05 respectively after a manual grid search in the space \{0.0001, 0.001, 0.05, 0.1, 0.5\} based on evaluating each generative metric based on the performance of original GCPN and GraphAF. We use Optuna to conduct a parallelised hyper-parameter grid search to determine the optimal hyper-parameters: the number of neurons in each hidden layer in the search space \{64, 128, 256\} and the learning rate for fine-tuning in the search space \{0.001, 0.0001, 0.00001, 0.000001\}. The number of neurons in each hidden layer and the learning rate for fine-tuning for each generative task are summarised below in Table \ref{tab:set1} and Table \ref{tab:set2}. For GraphEBM, we set the sample step of Langevin dynamics $K=150$, the standard deviation of the added noise in Langevin dynamics $\sigma=0.005$ and the step size $\lambda=30$ based on hyperparameters of the original energy-based model.

\begin{table}[ht]
	\centering
	\caption{\centering Detailed (batch size, learning rate) setup for GCPN and GraphAF variants on \newline Penalised logP, QED and SA.}
	\label{tab:set1}
	\begin{tabular}{lccc}
		\toprule
	\multirow{1}{*}{Model} & \multicolumn{1}{c}{Penalised logP} &                  \multicolumn{1}{c}{QED} & \multicolumn{1}{c}{SA} \\
		\midrule 
        GCPN (R-GCN) & 256, 0.00001 & 256, 0.00001 & 128, 0.001\phantom{0} \\
        GCPN (GIN) & 256, 0.0001\phantom{0}& 256, 0.00001&256, 0.0001 \\
        GCPN (GAT) & 128, 0.0001\phantom{0}& 256, 0.00001&\phantom{0}64, 0.001\phantom{0} \\
        GCPN (GATv2) & 256, 0.0001\phantom{0} & 256, 0.00001&256, 0.0001 \\
        GCPN (PNA) & 256, 0.00001&\phantom{0}64, 0.001\phantom{00}&256, 0.0001 \\
        GCPN (GSN) & 256, 0.0001\phantom{0}&\phantom{0}64, 0.001\phantom{00}&256, 0.0001 \\
        GCPN (GearNet) &256, 0.0001\phantom{0} & 256, 0.00001&128, 0.001\phantom{0} \\
        \midrule
        GraphAF (R-GCN) & \phantom{0}64, 0.0001\phantom{0} & 256, 0.00001 & 128, 0.0001 \\
        GraphAF (GearNet) & \phantom{0}64, 0.0001\phantom{0}&256, 0.000001 &128, 0.0001  \\
	    \bottomrule
        \end{tabular}
\end{table}

\begin{table}[ht]
	\centering
	\caption{\centering Detailed (batch size, learning rate) setup for GCPN, GraphAF and GraphEBM variants on DRD2, Median1 and Median2.}
	\label{tab:set2}
	\begin{tabular}{lccc}
		\toprule
	\multirow{1}{*}{Model} & \multicolumn{1}{c}{DRD2} & \multicolumn{1}{c}{Median1} & \multicolumn{1}{c}{Median2}\\
		\midrule 
        GCPN (R-GCN) & 256, 0.0001\phantom{00} & 128, 0.0001\phantom{00} & 256, 0.00001\phantom{0}\\
        GCPN (GIN) & 256, 0.00001\phantom{0} &256, 0.00001\phantom{0}& 256, 0.00001\phantom{0}\\
        GCPN (GAT) & 256, 0.00001\phantom{0}& \phantom{0}64, 0.001\phantom{000}& 256, 0.00001\phantom{0}\\
        GCPN (GATv2) & 128, 0.0001\phantom{00}&128, 0.001\phantom{000} & 256, 0.00001\phantom{0}\\
        GCPN (PNA) & \phantom{0}64, 0.001\phantom{000}& 256, 0.0001\phantom{00}& 256, 0.00001\phantom{0}\\
        GCPN (GSN) & 256, 0.00001\phantom{0}& \phantom{0}64, 0.001\phantom{000} & 256, 0.00001\phantom{0}\\
        GCPN (GearNet) & 256, 0.00001\phantom{0}& 256, 0.0001\phantom{00}& 256, 0.00001\phantom{0}\\
        \midrule
        GraphAF (R-GCN) & 256, 0.000001 &256, 0.00001\phantom{0} & 256, 0.000001 \\
        GraphAF (GIN) & 256, 0.000001 &256, 0.000001 &256, 0.000001 \\
        GraphAF (GAT) & 256, 0.000001 &256, 0.00001\phantom{0} &256, 0.00001\phantom{0} \\
        GraphAF (GATv2) & 256, 0.000001 &256, 0.000001 &256, 0.000001 \\
        GraphAF (PNA) & 256, 0.000001 &256, 0.000001 &256, 0.000001\\
        GraphAF (GSN) & 256, 0.0001\phantom{00} &256, 0.000001 &256, 0.000001\\
        GraphAF (GearNet) & 256, 0.00001\phantom{0}& 256, 0.00001\phantom{0}&256, 0.000001 \\
        \midrule
        GraphEBM (All) & 128, 0.0001\phantom{00} & 128, 0.0001\phantom{00} & 128, 0.0001\phantom{00} \\
	    \bottomrule
        \end{tabular}
\end{table}

\clearpage
\section{Appendix: Additional Results}
\label{meth}
\begin{figure}[h]
\centering 
\includegraphics[width=1\textwidth]{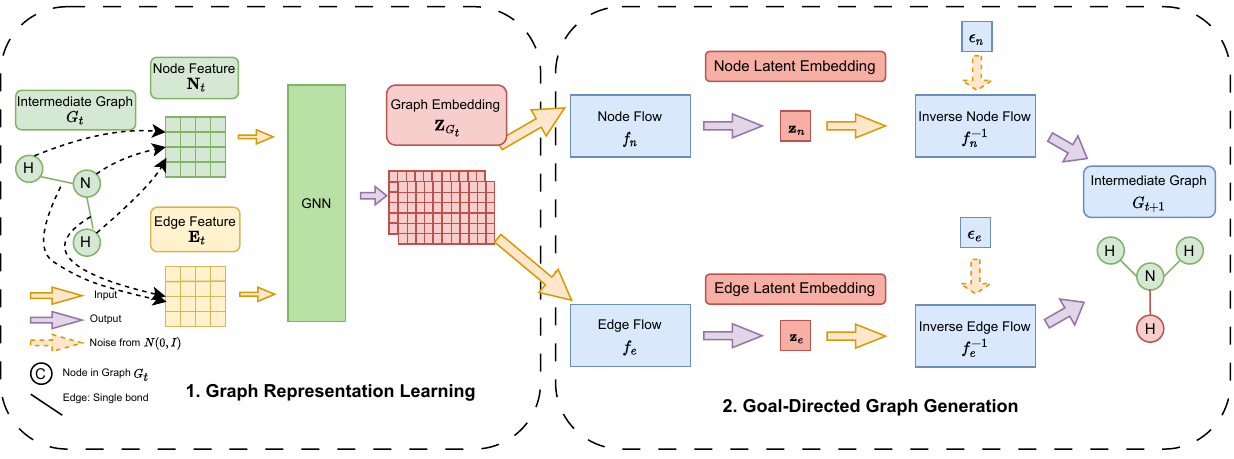}
\caption{An overview of the GraphAF model, demonstrating an example of iterative graph generation from an intermediate graph $G_{t}$ to an intermediate graph $G_{t+1}$. Part 1 is the illustration of the graph representation learning process based on a GNN. Part 2 is the illustration of the graph generative procedure based on a flow-based model. New nodes or edges are marked in red.}
\label{gnn2}
\end{figure}

\begin{figure}[h]
\centering 
\includegraphics[width=1\textwidth]{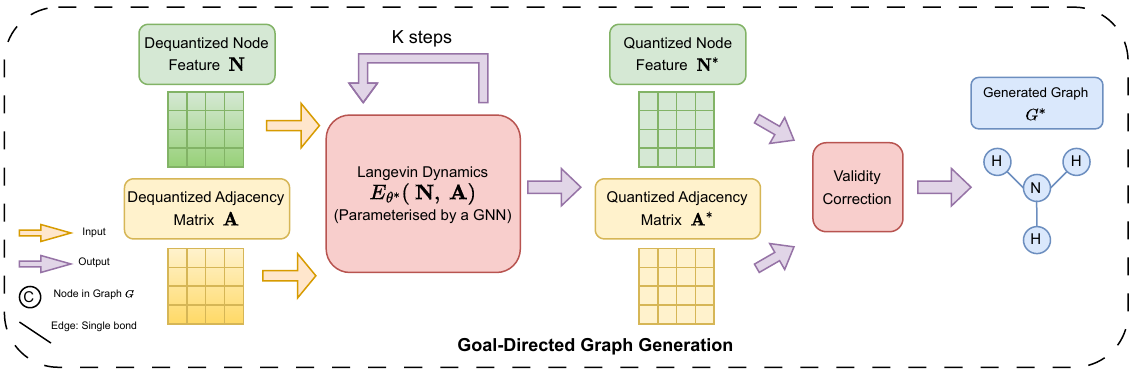}
\caption{An overview of the GraphEBM model, demonstrating an example of the one-shot graph generation from an initialised node feature matrix $\textbf{N}$ and adjacency matrix $\textbf{A}$ to a generated molecular graph $G^{*}$ with validity correction. It is the illustration of the graph generative procedure based on the Langevin dynamics with a trained energy function $E_{\theta^{*}}(\textbf{N}, \textbf{A})$ (parameterised by a GNN).}
\label{gnn3}
\end{figure}

\begin{table}[h!]
	\centering
	\caption{Comparison of the top-3 Penalised logP, QED and SA scores of the generated molecules by GCPN variants, with the top-3 property scores of molecules in the ZINC dataset for reference.}
	\label{tab:GCPN1}
	\begin{tabular}{lrrrcccccc}
		\toprule
        \multirow{2.5}{*}{Model} & \multicolumn{3}{c}{Penalised logP} & \multicolumn{3}{c}{QED} & \multicolumn{3}{c}{SA} \\
		\cmidrule(lr){2-4} \cmidrule(lr){5-7} \cmidrule(lr){8-10}
		& \(\mathbf{1st}\) & \(\mathbf{2nd}\) & \(\mathbf{3rd}\) & \(\mathbf{1st}\) & \(\mathbf{2nd}\) & \(\mathbf{3rd}\)& \(\mathbf{1st}\) & \(\mathbf{2nd}\) & \(\mathbf{3rd}\) \\
		\midrule
        ZINC & 4.52  & 4.30 & 4.23 & 0.948 & 0.948 & 0.948 & 1.0 & 1.0 & 1.0\\
        \midrule
        R-GCN (baseline) & 7.98 & 7.85 & 7.80 & 0.948 & 0.947 & 0.946 & --- & --- & ---\\
		R-GCN (ours) & 8.67  & 8.67 & 8.67 & 0.948 & 0.948 & 0.948 & \color{DarkGreen}\textbf{1.0} & \color{DarkGreen}\textbf{1.0} & \color{DarkGreen}\textbf{1.0}\\
		GIN & \color{DarkGreen}\textbf{11.19}  & \color{DarkGreen}\textbf{11.19} & \color{DarkGreen}\textbf{11.19} & 0.942 & 0.926 & 0.923 & 1.2 & 1.2 & 1.2\\
		GAT & 7.70  & 7.47 & 7.44 & 0.926 & 0.911 & 0.911 & \color{DarkGreen}\textbf{1.0} & \color{DarkGreen}\textbf{1.0} & \color{DarkGreen}\textbf{1.0}\\
        GATv2 & 8.08  & 7.48 & 7.34 & 0.945 & 0.908 & 0.907 & \color{DarkGreen}\textbf{1.0} & \color{DarkGreen}\textbf{1.0} & \color{DarkGreen}\textbf{1.0} \\
        PNA & 8.66  & 8.61 & 8.22 & 0.833 & 0.825 & 0.754 & \color{DarkGreen}\textbf{1.0} & \color{DarkGreen}\textbf{1.0} & \color{DarkGreen}\textbf{1.0}\\
        GSN & \color{DarkGreen}\textbf{11.19}  & \color{DarkGreen}\textbf{11.19} & \color{DarkGreen}\textbf{11.19}  & 0.804 & 0.783 & 0.783 & \color{DarkGreen}\textbf{1.0} & \color{DarkGreen}\textbf{1.0} & \color{DarkGreen}\textbf{1.0}\\
        GearNet & \color{DarkGreen}\textbf{11.19}  & \color{DarkGreen}\textbf{11.19} & \color{DarkGreen}\textbf{11.19} & \color{DarkGreen}\textbf{0.948} & \color{DarkGreen}\textbf{0.948} & \color{DarkGreen}\textbf{0.948} & \color{DarkGreen}\textbf{1.0} & \color{DarkGreen}\textbf{1.0} & \color{DarkGreen}\textbf{1.0}\\
		\bottomrule
	\end{tabular}
\end{table}

\begin{table}[ht]
	\centering
	\caption{Comparison of the top-3 DRD2, Median1 and Median2 scores of the generated molecules by GCPN--e variants, with the top-3 property scores of molecules in the ZINC dataset for reference.}
	\label{tab:GCPN3}
  \resizebox{\textwidth}{!}{
	\begin{tabular}{lccccccccc}
		\toprule
	\multirow{2.5}{*}{Model} & \multicolumn{3}{c}{DRD2} & \multicolumn{3}{c}{Median1} & \multicolumn{3}{c}{Median2}\\
		\cmidrule(lr){2-4} \cmidrule(lr){5-7} \cmidrule(lr){8-10}
		& \(\mathbf{1st}\) & \(\mathbf{2nd}\) & \(\mathbf{3rd}\) & \(\mathbf{1st}\) & \(\mathbf{2nd}\) & \(\mathbf{3rd}\) & \(\mathbf{1st}\) & \(\mathbf{2nd}\) & \(\mathbf{3rd}\)\\
		\midrule
            ZINC & 0.9872  & 0.9815 & 0.9773 & 0.3243 & 0.3096 & 0.3096 & 0.2913 & 0.2765 & 0.2749 \\
            \midrule
		R-GCN & 0.8315  & 0.7576 & 0.7551 &0.3152&0.3152&0.3001 & 0.1932 & 0.1613 & 0.1592 \\
		GIN & 0.2791  & 0.1980 & 0.1752 &0.3152&0.3152&0.3152& 0.1140 & 0.1113 & 0.1069\\
		GAT  & 0.1980 & 0.1580 & 0.1580 & 0.3243&0.3243&0.3175 & 0.1196 & 0.1100  & 0.1098 \\
            GATv2 &0.2992  & 0.2992 & 0.2992 & 0.3281&0.3281&0.3281 & 0.1042 & 0.1009 & 0.0911\\
            PNA & 0.4448  & 0.4448 & 0.4448 &0.3243&0.3202&0.3175 & 0.0911 & 0.0764 & 0.0716\\
            GSN & 0.4790  & 0.4790 & 0.4448 &0.3175&0.3175&0.3015& 0.0982 & 0.0978 & 0.0897\\
            GearNet & \color{DarkGreen}\textbf{0.9990}  & \color{DarkGreen}\textbf{0.9705} & \color{DarkGreen}\textbf{0.9574} &\color{DarkGreen}\textbf{0.3482}&\color{DarkGreen}\textbf{0.3482}&\color{DarkGreen}\textbf{0.3482} & \color{DarkGreen}\textbf{0.2084} & \color{DarkGreen}\textbf{0.2043} & \color{DarkGreen}\textbf{0.2037} \\
		\bottomrule
	\end{tabular}}
\end{table}

\begin{table}[ht]
	\centering
	\caption{Comparison of the top-3 DRD2, Median1 and Median2 scores of the generated molecules by GraphAF variants, with the top-3 property scores of molecules in the ZINC dataset for reference.}
	\label{tab:GraphAF1}
 \resizebox{\textwidth}{!}{
	\begin{tabular}{lccccccccc}
		\toprule
	\multirow{2.5}{*}{Model} & \multicolumn{3}{c}{DRD2} & \multicolumn{3}{c}{Median1} & \multicolumn{3}{c}{Median2}\\
		\cmidrule(lr){2-4} \cmidrule(lr){5-7} \cmidrule(lr){8-10}
		& \(\mathbf{1st}\) & \(\mathbf{2nd}\) & \(\mathbf{3rd}\) & \(\mathbf{1st}\) & \(\mathbf{2nd}\) & \(\mathbf{3rd}\) & \(\mathbf{1st}\) & \(\mathbf{2nd}\) & \(\mathbf{3rd}\)\\
            \midrule
            ZINC & 0.9872  & 0.9815 & 0.9773 & 0.3243 & 0.3096 & 0.3096 & 0.2913 & 0.2765 & 0.2749 \\
            \midrule
		R-GCN & 0.9277 & 0.9133 & 0.9080 &0.2810&0.2641&0.2449 & 0.1426 & 0.1426 & 0.1417 \\
            GIN & 0.5847&0.1835&0.1495&0.2651&0.2649&0.2393& 0.1094 &0.1042 & 0.1037 \\
		GAT & 0.2992  & 0.2992 & 0.2992&0.2453&0.2212&0.2208 & 0.1031 & 0.1025 & 0.1023 \\
            GATv2 & 0.5909  & 0.4790 & 0.4790 & 0.2437 & 0.2335  & 0.2331 & 0.1239 & 0.1239 & 0.1232\\
            PNA & 0.1495  & 0.1411 & 0.1411 & 0.2897 & 0.2651 & 0.2651 & 0.1023 & 0.0764 & 0.0761\\
            GSN & 0.1238  & 0.0614 & 0.0601 & 0.2896 & 0.2773 & 0.2449 & 0.1025 & 0.1017 & 0.0977\\
            GearNet & \color{DarkGreen}\textbf{0.9872}  & \color{DarkGreen}\textbf{0.9725} & \color{DarkGreen}\textbf{0.9714} &\color{DarkGreen}\textbf{0.2897} & \color{DarkGreen}\textbf{0.2896} & \color{DarkGreen}\textbf{0.2651} & \color{DarkGreen}\textbf{0.1826} & \color{DarkGreen}\textbf{0.1666} & \color{DarkGreen}\textbf{0.1616}\\
		\bottomrule
	\end{tabular}}
\end{table}

\begin{table}[ht]
	\centering
	\caption{Comparison of the top-3 DRD2, Median1 and Median2 scores of the generated molecules by GraphAF+e variants, with the top-3 property scores of molecules in the ZINC dataset for reference.}
	\label{tab:GraphAF2}
 \resizebox{\textwidth}{!}{
	\begin{tabular}{lccccccccc}
		\toprule
	\multirow{2.5}{*}{Model} & \multicolumn{3}{c}{DRD2} & \multicolumn{3}{c}{Median1} & \multicolumn{3}{c}{Median2}\\
		\cmidrule(lr){2-4} \cmidrule(lr){5-7} \cmidrule(lr){8-10}
		& \(\mathbf{1st}\) & \(\mathbf{2nd}\) & \(\mathbf{3rd}\) & \(\mathbf{1st}\) & \(\mathbf{2nd}\) & \(\mathbf{3rd}\) & \(\mathbf{1st}\) & \(\mathbf{2nd}\) & \(\mathbf{3rd}\)\\
  		\midrule
            ZINC & 0.9872  & 0.9815 & 0.9773 & 0.3243 & 0.3096 & 0.3096 & 0.2913 & 0.2765 & 0.2749 \\
            \midrule
		R-GCN & 0.7833  & 0.7702 & 0.5714 & {\color{DarkGreen}\textbf{0.2651}} & {\color{DarkGreen}\textbf{0.2667}} & {\color{DarkGreen}\textbf{0.2632}} & 0.1526 & 0.1507 & 0.1481 \\
		GIN & 0.8391  & 0.8016 & 0.7145 & 0.2735 & 0.2343 & 0.2304 & 0.1505 & 0.1496 & 0.1459 \\
		GAT & 0.4790  & 0.4790 & 0.4790 & 0.2651 & 0.2509 & 0.2471 & 0.1358 & 0.1328 & 0.1152 \\
            GATv2 & 0.7087  & 0.5452 & 0.3986 & 0.2887 & 0.2342  & 0.2331 & 0.1362 & 0.1343 & 0.1316\\
            PNA & 0.4940  & 0.4940 & 0.4940 & 0.2633 & 0.2633 & 0.2449 & 0.1423 & 0.1313 & 0.1110\\
            GSN & 0.5106  & 0.4940 & 0.4940 & 0.2572 & 0.2530 & 0.2518 & 0.1448 & 0.1420 & 0.1090\\
            GearNet & \color{DarkGreen}\textbf{0.9699}  & \color{DarkGreen}\textbf{0.9688} & \color{DarkGreen}\textbf{0.9623} & 0.2810 & 0.2582 & 0.2453 & \color{DarkGreen}\textbf{0.1829}& \color{DarkGreen}\textbf{0.1795} & \color{DarkGreen}\textbf{0.1672}\\
		\bottomrule
	\end{tabular}}
\end{table}

\begin{table}[ht]
	\centering
	\caption{Comparison of the top-3 DRD2, Median1 and Median2 scores of the generated molecules by GraphEBM variants, with the top-3 property scores of molecules in the ZINC dataset for reference.}
	\label{tab:GraphEBM}
 \resizebox{\textwidth}{!}{
	\begin{tabular}{lccccccccc}
		\toprule
	\multirow{2.5}{*}{Model} & \multicolumn{3}{c}{DRD2} & \multicolumn{3}{c}{Median1} & \multicolumn{3}{c}{Median2}\\
		\cmidrule(lr){2-4} \cmidrule(lr){5-7} \cmidrule(lr){8-10}
		& \(\mathbf{1st}\) & \(\mathbf{2nd}\) & \(\mathbf{3rd}\) & \(\mathbf{1st}\) & \(\mathbf{2nd}\) & \(\mathbf{3rd}\) & \(\mathbf{1st}\) & \(\mathbf{2nd}\) & \(\mathbf{3rd}\)\\
  		\midrule
            ZINC & 0.9872  & 0.9815 & 0.9773 & 0.3243 & 0.3096 & 0.3096 & 0.2913 & 0.2765 & 0.2749 \\
            \midrule
		R-GCN & 0.6907  & 0.6907 & 0.6907 & 0.2810 & 0.2810 & 0.2810 & 0.1481 & 0.1481 & 0.1481 \\
		GIN & 0.6907  & 0.6907 & 0.6907 & 0.2357 & 0.2357 & 0.2357 & 0.1420 & 0.1420 & 0.1420 \\
		GAT & 0.6907  & 0.6907 & 0.6907 & 0.2897 & 0.2897 & 0.2897 & 0.1313 & 0.1313 & 0.1313 \\
            GATv2 & 0.8305  & 0.8305 & 0.8305 & 0.3243 & 0.3243 & 0.3243 & 0.1640 & 0.1640 & 0.1640\\
            PNA & 0.7652  & 0.7652 & 0.7652 & 0.2053 & 0.2053 & 0.2053 & 0.1343 & 0.1343 & 0.1343\\
            GSN & 0.7188  & 0.7188 & 0.7188 & 0.2810 & 0.2810 & 0.2810 & 0.1672 & 0.1672 & 0.1672\\
            GearNet & \color{DarkGreen}\textbf{0.9443}  & \color{DarkGreen}\textbf{0.9443} & \color{DarkGreen}\textbf{0.9443} & {\color{DarkGreen}\textbf{0.3999}} & {\color{DarkGreen}\textbf{0.3999}} & {\color{DarkGreen}\textbf{0.3999}} & \color{DarkGreen}\textbf{0.2067}& \color{DarkGreen}\textbf{0.2067} & \color{DarkGreen}\textbf{0.2067}\\
		\bottomrule
	\end{tabular}}
\end{table}

\begin{table}[ht]
	\centering
	\caption{Comparison of the top-1 DRD2, Median1, Median2 and QED scores between the experimented GNN-based generative model variants (GCPN, GraphAF, and GraphEBM) and all non-GNN-based graph generative models in the benchmark~\cite{benchmark}.}
	\label{tab:generative2}
        \resizebox{\textwidth}{!}{
	\begin{tabular}{lllll}
		\toprule
	\multirow{1}{*}{Model} & \multicolumn{1}{l}{DRD2} & \multicolumn{1}{l}{Median1} & \multicolumn{1}{l}{Median2} & \multicolumn{1}{l}{QED}\\
             \midrule
            GCPN (R-GCN) & 0.479 & 0.337 & 0.192 & 0.948\\
            GCPN (GearNet) & 0.970 \textbf{($+$102.51$\%$)} & 0.337 \textbf{($+$0.00$\%$)} & 0.286 \textbf{($+$48.96$\%$)} & 0.948 \textbf{($+$0.00$\%$)}\\
            \midrule
            GraphAF (R-GCN) & 0.928 & 0.281 & 0.143 & 0.946\\
            GraphAF (GearNet) & 0.987 \textbf{($+$6.36$\%$)} & 0.290 \textbf{($+$3.20$\%$)} & 0.183 \textbf{($+$27.97$\%$)} & 0.947 \textbf{($+$0.11$\%$)}\\
            \midrule
            GraphEBM (R-GCN) & 0.691 & 0.281 & 0.148 & 0.948\\
            GraphEBM (GearNet) & 0.944 \textbf{($+$36.61$\%$)} & 0.400 
            \textbf{($+$42.35$\%$)} & 0.207 \textbf{($+$39.86$\%$)} & 0.948 \textbf{($+$0.00$\%$)}\\
            \midrule
            REINVENT (SMILES)~\cite{drd2} & 0.999 & 0.399 &0.332 & 0.948\\
            LSTM HC (SMILES)~\cite{dataset1} & 0.999 & 0.388 & 0.339 & 0.948 \\
            Graph GA~\cite{Graph-MCTS} & 0.999& 0.350& 0.324&0.948\\
            REINVENT (SELFIES)~\cite{drd2} & 0.999 & 0.399 & 0.313 & 0.948\\
            DoG-Gen~\cite{DoG-Gen} & 0.999 & 0.322 & 0.297 & 0.948 \\
            GP BO~\cite{GPBO} & 0.999 & 0.345 & 0.337 & 0.947 \\
            STONED~\cite{sto} & 0.997 & 0.295& 0.265& 0.947 \\
            LSTM HC (SELFIES)~\cite{dataset1} & 0.999 & 0.362& 0.274& 0.948\\
            DST~\cite{dst} &0.999 &0.281 &0.201 &0.947 \\
            SMILES GA~\cite{dataset1} & 0.986& 0.207&0.210 & 0.948\\
            SynNet~\cite{SynNet} & 0.999 & 0.244 & 0.259 & 0.948 \\
            MARS~\cite{mars} & 0.994& 0.233&0.203 &0.946 \\
            MolPal~\cite{molpal} &0.964 &0.309 & 0.273& 0.948\\
            MIMOSA~\cite{mimosa} & 0.993& 0.296& 0.238& 0.947\\
            GA+D~\cite{GA+D} & 0.836 & 0.219 & 0.161 & 0.945 \\
            VAE BO (SELFIES)~\cite{SMILES-VAE} & 0.940& 0.231& 0.206& 0.947\\
            DoG-AE~\cite{DoG-Gen} &0.999 &0.203 &0.201 & 0.944\\
            Screening~\cite{screen} &0.949 &0.271 &0.244 &0.947 \\
            GFlowNet~\cite{gflow} &0.951 & 0.237&0.198 & 0.945\\
            VAE BO (SMILES)~\cite{SMILES-VAE} & 0.940 & 0.231 & 0.206 & 0.947 \\
            Pasithea~\cite{pas} & 0.592&0.216 &0.194 &0.943 \\
            JT-VAE BO~\cite{jt} &0.778 &0.212 &0.192 &0.946 \\
            GFlowNet-AL~\cite{gflow} &0.863 &0.229 &0.191 &0.944 \\
            Graph MCTS~\cite{Graph-MCTS} & 0.586 & 0.242 & 0.148 & 0.928 \\
            MolDQN~\cite{MolDQN} & 0.049 & 0.188 & 0.108 & 0.871 \\
		\bottomrule
	\end{tabular}}
\end{table}

\clearpage
\section{Appendix: Visualisation of Generated Molecule Graphs}
\label{appb}

We present visualisations of generated molecules with the highest score on metric Penalised logP and QED. The visualisations illustrate that generated molecules with the highest Penalised logP score only contain a long chain of carbons. 

\begin{figure}[h]
\centering 
\subfigure[Generated molecules with the highest Penalised logP scores]{
\label{c5}
\includegraphics[width=0.6\textwidth]{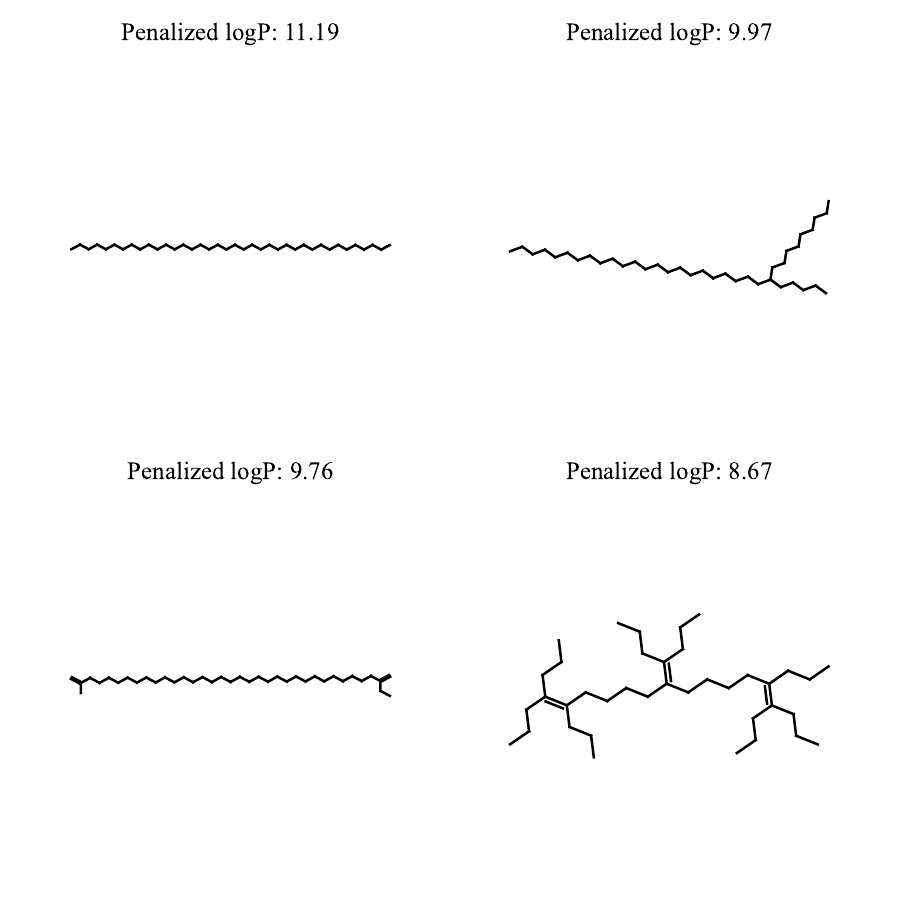}}
\subfigure[Generated molecules with the highest QED scores]{
\label{c6} 
\includegraphics[width=0.6\textwidth]{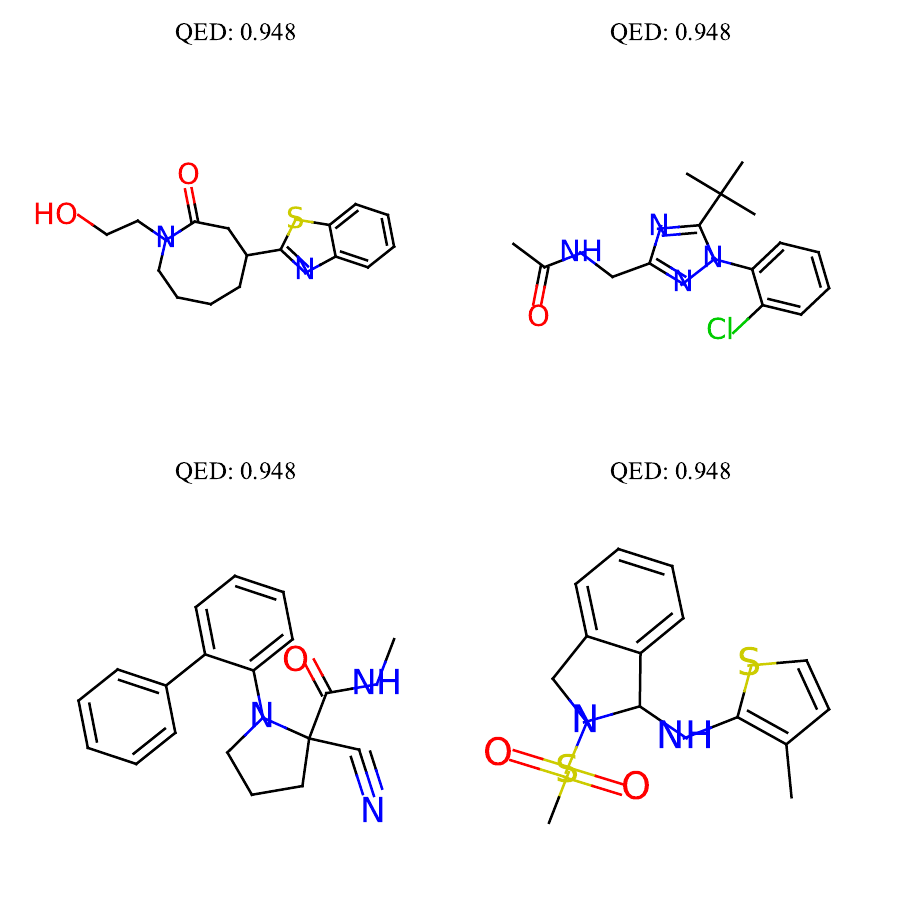}}\\

\caption{Molecules with highest generation metrics: Penalised logP and QED generated by GNN-based graph generative models on de-novo molecule design tasks.}
\label{d}
\end{figure}

\begin{figure}[h]
\centering 
\subfigure[Molecules with most desired generation metrics generated by GCPN (R-GCN)]{
\label{c1}
\includegraphics[width=0.45\textwidth]{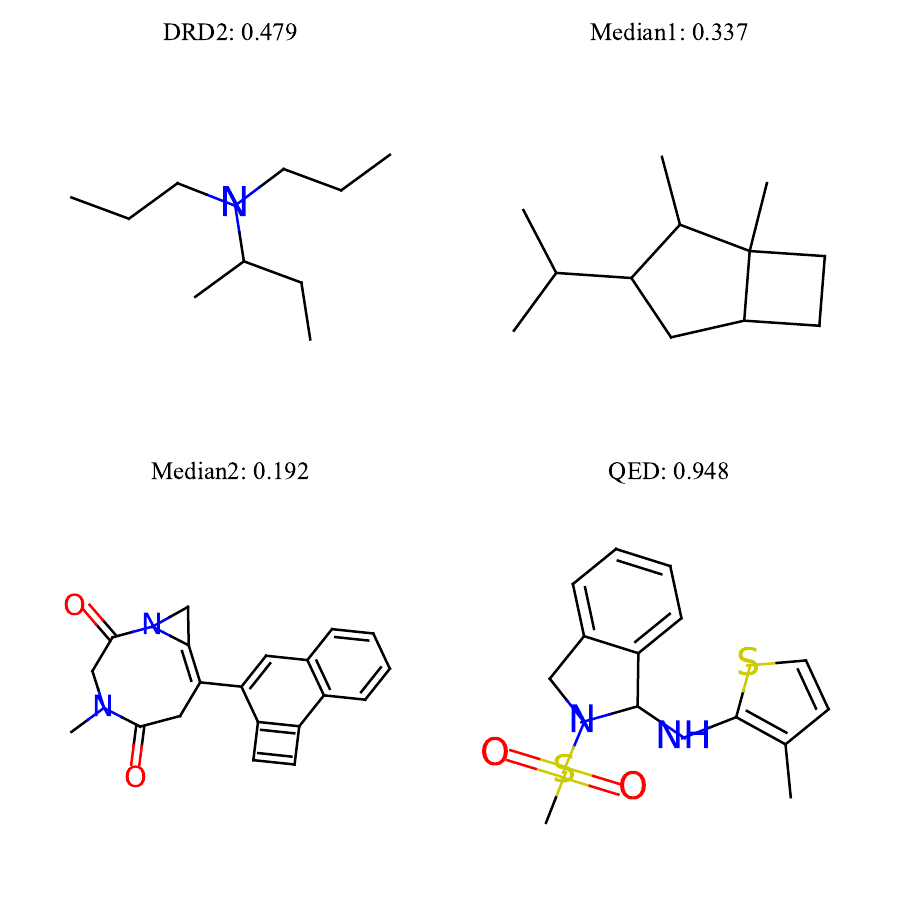}}
\hfill
\subfigure[Molecules with most desired generation metrics generated by GraphAF (R-GCN)]{
\label{c2} 
\includegraphics[width=0.45\textwidth]{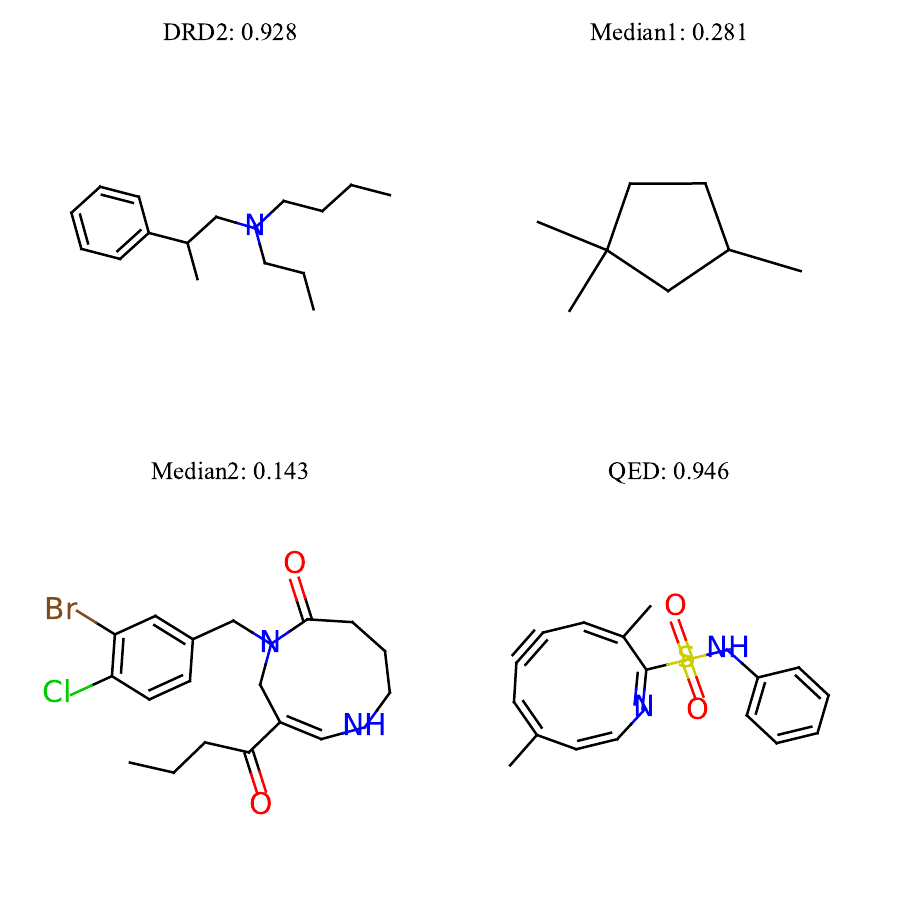}}\\
\subfigure[Molecules with most desired generation metrics generated by GCPN (GearNet)]{
\label{c3}
\includegraphics[width=0.45\textwidth]{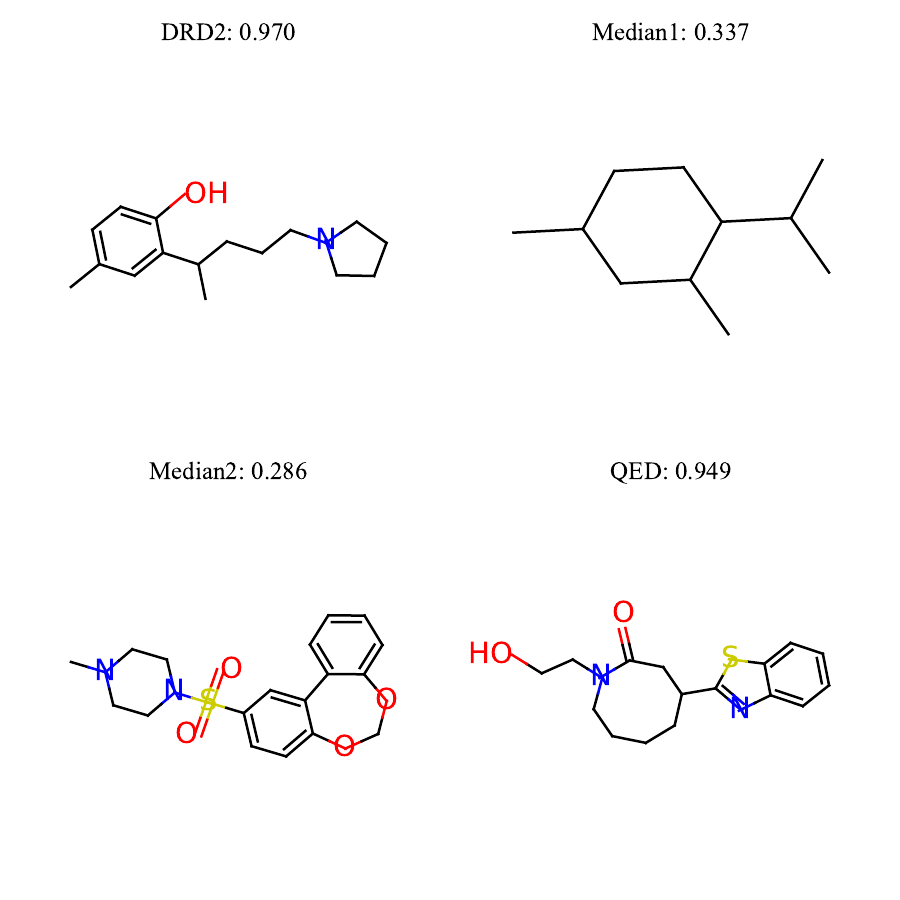}}
\hfill
\subfigure[Molecules with most desired generation metrics generated by GraphAF (GearNet)]{
\label{c4}
\includegraphics[width=0.45\textwidth]{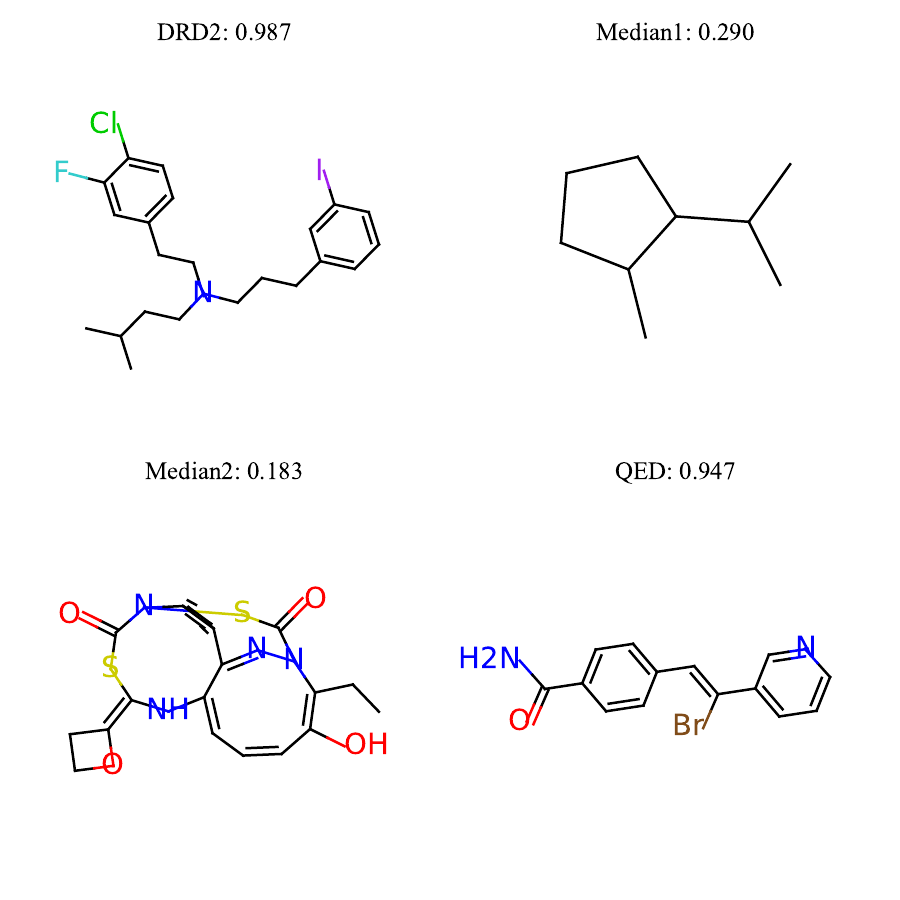}}\\

\caption{Molecules with highest generation metrics: DRD2, Median1, Median2 and QED generated by proposed GNN-based graph generative models: (a) GCPN with R-GCN (b) GraphAF with R-GCN (c) GCPN with GearNet (d) GraphAF with GearNet on de-novo molecule design tasks.}
\label{c}
\end{figure}

\clearpage
We present visualisations of generated molecules by GCPN, GraphAF and their optimal variants with GearNet in Figure \ref{c1}, Figure \ref{c2}, Figure \ref{c3} and Figure \ref{c4} respectively. The visualisations demonstrate that GCPN and GraphAF with advanced GNN have strong abilities to model different graph structures on the de-novo molecule design task.

\end{document}